\documentclass[]{article}

\PassOptionsToPackage{numbers, compress}{natbib}

\usepackage[preprint]{neurips_2024} 

\usepackage[utf8]{inputenc} 
\usepackage[T1]{fontenc}    
\usepackage{hyperref}       
\usepackage{url}            
\usepackage{booktabs}       
\usepackage{amsfonts}       
\usepackage{nicefrac}       
\usepackage{microtype}      
\usepackage{tablefootnote}
\usepackage{comment}

\usepackage{etoolbox} 
\usepackage{amsmath}
\usepackage{amssymb}
\usepackage{graphicx}
\usepackage{subcaption}
\usepackage{listings}
\usepackage{float}
\usepackage{enumitem}
\usepackage{xspace}
\usepackage{multirow}
\usepackage{array}
\usepackage{tabularx}
\usepackage{booktabs}
\usepackage{xcolor,colortbl}
\usepackage{cleveref}
\usepackage[english]{babel}
\usepackage{pifont} 
\usepackage{soul}
\usepackage{slashbox}
\usepackage{xfrac}
\usepackage{wrapfig}
\usepackage{mdwlist}
\usepackage{arydshln}
\usepackage[obeyFinal,disable]{todonotes}

\usepackage{adjustbox}

\usepackage[ruled,vlined]{algorithm2e}

\newtheorem{definition}{Definition}

\newtheorem{theorem}{Theorem}

\newtheorem{proof}{Proof}

\renewcommand{\paragraph}[1]{\noindent\textbf{#1}.}

\crefformat{section}{\S#2#1#3} 
\crefformat{subsection}{\S#2#1#3}
\crefformat{subsubsection}{\S#2#1#3}
\newcolumntype{L}[1]{>{\raggedright\let\newline\\\arraybackslash\hspace{0pt}}m{#1}}
\newcolumntype{C}[1]{>{\centering\let\newline\\\arraybackslash\hspace{0pt}}m{#1}}
\newcolumntype{R}[1]{>{\raggedleft\let\newline\\\arraybackslash\hspace{0pt}}m{#1}}

\SetCommentSty{mycommfont}

\newcommand{\jiong}[1]{\todo[color=orange]{JoZ: #1}}

\definecolor{LightCyan}{rgb}{0.88,1,1}
\definecolor{purple2}{RGB}{153,0,153} 
\definecolor{green2}{RGB}{0,153,0} 

\definecolor{darkgreen}{RGB}{0,153,0}

\lstdefinestyle{common}{
	basicstyle = \ttfamily,
	keywordstyle=\color{blue},       
	keywordstyle={[2]\color{cyan}}, 
	stringstyle=\color{purple2},
	commentstyle=\color{green2},
	upquote=true,                  
	breaklines=true, frame=trBL
}

\newlist{steps}{enumerate}{1}
\setlist[steps, 1]{label = Step \arabic*:}

\newcommand{\V}[1]{\mathbf{#1}}
\newcommand{\R}{\mathbb{R}}

\newcommand{\graph}{\mathcal{G}}
\newcommand{\vertexSet}{\mathcal{V}}
\newcommand{\edgeSet}{\mathcal{E}}

\newcommand{\matA}{\mathbf{A}}

\newcommand{\matX}{\mathbf{X}}

\newcommand{\T}{\mathsf{T}}

\newcommand{\neighNoSelfLoop}{\bar{N}}

\newcolumntype{H}{>{\setbox0=\hbox\bgroup}c<{\egroup}@{}}

\makeatletter
\newcommand\footnoteref[1]{\protected@xdef\@thefnmark{\ref{#1}}\@footnotemark}
\makeatother
\title{On the Impact of Feature Heterophily on Link Prediction with Graph Neural Networks}

\author{
  Jiong Zhu\thanks{Equal contribution.} \\
  University of Michigan \\
  \texttt{jiongzhu@umich.edu} \\
  \And
  Gaotang Li\footnotemark[1] \\
  University of Illinois\\
Urbana-Champaign \\
  \texttt{gaotang3@illinois.edu} \\
  \And
  Yao-An Yang \\
  University of Michigan \\
  \texttt{ayayang@umich.edu} \\
  \And
  Jing Zhu \\
  University of Michigan\\
  \texttt{jingzhuu@umich.edu} \\
  \And
  Xuehao Cui \\
  University of Michigan \\
  \texttt{credus@umich.edu} \\
  \And
  Danai Koutra \\
  University of Michigan\\
  \texttt{dkoutra@umich.edu}
}

\begin{document}


\maketitle

\begin{abstract}
    Heterophily, or the tendency of connected nodes in networks to have different class labels or dissimilar features, has been identified as challenging for many Graph Neural Network (GNN) models.
    While the challenges of applying GNNs for node classification when \emph{class labels} display strong heterophily are well understood, it is unclear how heterophily 
    affects GNN performance in 
    other important graph learning tasks where class labels are not available.
    In this work, we focus on the link prediction task and systematically analyze the impact of heterophily in \textit{node features} on GNN performance. Theoretically, we first introduce formal definitions of homophilic and heterophilic link prediction tasks,   
    and present a theoretical framework that highlights the different optimizations needed for the respective tasks. 
    We then analyze how different link prediction encoders and decoders adapt to varying levels of feature homophily and introduce designs for improved performance. 
    Our empirical analysis on a variety of synthetic and real-world datasets confirms our theoretical insights and highlights the importance of adopting learnable decoders and GNN encoders with ego- and neighbor-embedding separation in message passing for link prediction tasks beyond homophily. 

\end{abstract}

\section{Introduction}

Graph-structured data are powerful and widely used in the real world, representing relationships beyond those in Euclidean data through links. Link prediction, which aims to predict missing edges in a graph, is an important task. It has fundamental applications in recommendation systems~\cite{vahidi2021hybrid}, knowledge graphs~\cite{rossi2021knowledge}, and social networks~\cite{daud2020applications}. Traditional algorithms for link prediction are heuristic-based and impose strong assumptions on the link generation process. To alleviate the reliance on handcrafted features, recent link prediction approaches are transformed by Graph Neural Networks (GNNs), which can effectively learn node representations in an end-to-end fashion. Vanilla GNN for link prediction (GNN4LP) methods keep the original GNN model for encoding node embeddings, followed by a decoder acting on pairwise node embeddings, \emph{e.g.} dot product~\cite{kipf2016variational}. However, these methods are not good at capturing pairwise structural proximity information~\cite{zhang2021labeling,liang2022can}, \emph{i.e.} neighborhood heuristics such as the number of common neighbors. 
To further enhance model capabilities, the current \emph{state-of-the-art} GNN4LP approaches augment GNNs by incorporating pairwise structural information~\cite{chamberlain2022graph-ct, zhang2018link, zhang2022graph,yun2022neo-gnns-cr,wang2023neural}. 

Nevertheless, as the core of today's \emph{SOTA} approaches, GNNs rely on the \emph{message-passing} mechanism
and are naturally built on the homophily assumption, \emph{i.e.} connected nodes tend to share similar attributes with each other. Such inductive bias has been widely analyzed and has been shown to be an important factor for GNN's superior performance in the task of node classification on homophilic graphs~\cite{zheng2022graph,halcrow2020grale,luan2023when,ma2022is}. It has also been widely observed that GNN's performance degrades on heterophilic graphs in node classification tasks, where connected nodes tend to have different labels~\cite{abu2019mixhop,zhu2020graph, zhu2020beyond,pei2020geom,he2021bernnet,luan2022revisiting}.
In contrast, there are merely works paying attention to the problem of heterophily in the link prediction settings: {almost all existing definitions of homophily rely on the node class labels~\cite{lim2021new,platonov2024characterizing}, which are often not available for the link prediction tasks.
Furthermore, prior works on GNN4LP have largely focused on the effects of pairwise structural information to link prediction performance, while there is no dedicated work focusing on the effects of feature heterophily.} 
In light of this, this work aims to characterize a notion of heterophily in link prediction problem, understand the effects of heterophily and feature similarity in existing models that leverage node features, and explore designs to improve the use of dissimilar features in GNN link prediction. We detail our contributions as follows:

\begin{itemize}[leftmargin=*]
    \item \textbf{Definitions of Non-homophilic Link Prediction}: We introduce formal definitions of homophilic and non-homophilic link prediction tasks: instead of relying on the magnitude of feature similarity, our definitions are based on the separation of feature similarity scores between edges and non-edges, which is justified by a concise theoretical framework that highlights the different optimizations needed for the respective tasks.
    
    \item \textbf{Designs Empowering GNNs for Non-homophilic Link Prediction}: We identify designs for GNN encoders and link probability decoders that improve performance for non-homophilic link prediction settings
    and show that (1) decoders with sufficient complexity are required for capturing non-homophilic feature correlations between connected nodes; 
    (2) ego- and neighbor-embedding separation in GNN message passing improves their adaptability to feature similarity variations. 
    
    \item \textbf{Empirical Analysis on Impacts of Feature Heterophily}: We create synthetic graphs  with varying levels of feature similarity and compare the performance trend of various link prediction methods with different designs for non-homophilic link prediction. On real-world graphs, we compare the overall link prediction performance of different methods on datasets with diverse feature similarities; in addition, we further conduct a zoom-in analysis for link prediction performance on different group of edges bucketized by node degrees and feature similarity scores.  
\end{itemize}
\section{Notation and Preliminaries} 
\label{sec:preliminaries}

Let $\graph=(\vertexSet,\edgeSet)$ be an undirected and unweighted homogeneous graph with node set $\vertexSet$ and edge set $\edgeSet$. 
We denote the 1-hop (immediate) neighborhood centered around $v$ as $N(v)$ ($\graph$ may have self-loops), and the corresponding neighborhood that does \textit{not} include the ego (node $v$) as $\neighNoSelfLoop(v)$.
We represent the graph by 
its adjacency matrix $\matA \in \{0,1\}^{n\times n}$ 
and its node feature set as $\mathcal{X}$ with matrix form $\matX \in \mathbb{R}^{n \times F}$, where 
the vector $\V{x}_v$ corresponds to the \textit{ego-feature} of node $v$, 
and $\{\V{x}_u: u \in \neighNoSelfLoop(v)\}$ to its \textit{neighbor-features}. 
We further represent the degree of a node $v$ by $d_v$, which denotes the number of neighbors in its immediate neighborhood $\neighNoSelfLoop(v)$.

\paragraph{Graph Neural Networks for Link Prediction} Following \cite{li2023evaluating-lb,zhang2020revisiting}, we define the task of link prediction to be estimating the likelihood of reconstructing the actual adjacency matrix. Formally, 
\begin{equation}
    \hat{y}_{i,j} = \hat{\matA}_{i,j} = p(i, j | \graph, \matX)
\end{equation}

where $\hat{y}_{i,j}$ or $\hat{\matA}_{i,j}$ is the predicted link probability between nodes $(i,j)$ and was traditionally calculated by heuristics-based algorithms. 
For $(i, j)$ in the training set, we set the ground truth probability $y_{i,j} = \matA_{i,j}$.
Existing GNN4LP approaches typically use a GNN-based method for encoding node representations (denoted by ENC) and some decoder function (denoted by DEC) between node embedding pairs:

\begin{equation}
    \hat{y}_{i,j} = \hat{\matA}_{i, j} = \text{DEC}(z_i, z_j), \text{ where } z_i = \text{ENC}(i, \graph, \matX), z_j = \text{ENC}(j, \graph, \matX)  
\end{equation}

The original graph autoencoder approach \cite{kipf2016variational} uses a two-layer GCN as the encoder and a dot product as the decoder. There are many different choices of encoders that do not need to strictly follow the original GNN architecture. For the decoder, a popular simple choice is the dot product, but one can use a set function such as concatenation followed by an MLP as well. 

\paragraph{Graph Feature Similarity} We measure the ``graph feature similarity'' through averaging the \emph{mean-centered} feature similarity from connected node pairs. 

\begin{definition}[Node Feature Similarity]
    For a node pair $(u, v)$ with $\mathbf{x}_u$ and $\mathbf{x}_v$ as the node features and a similarity function $\phi \colon (\cdot, \cdot) \mapsto \R$, we define the node feature similarity as $k(u, v) = \phi(\mathbf{x}_u, \mathbf{x}_v)$.
\end{definition}

In this work, we set $\phi(\mathbf{x}_u, \mathbf{x}_v) = \tfrac{\bar{\mathbf{x}}_u \cdot \bar{\mathbf{x}}_v}{\Vert\bar{\mathbf{x}}_u\Vert \Vert\bar{\mathbf{x}}_v\Vert}$
to be the \emph{mean-centered} cosine similarity of node features.
Specifically, we denote the mean feature vector of all nodes in the graph as $\bar{\mathbf{X}} = \tfrac{1}{|\vertexSet|} \sum_{v \in \vertexSet} \mathbf{x}_v$, and the mean-centered node feature for node $v$ as $\bar{\mathbf{x}}_v = \mathbf{x}_v - \bar{\mathbf{X}}$. Empirically, we find that the mean-centering operation is crucial for accurate characterization of the pairwise feature similarity and their impact on link prediction performance.
We further define the graph feature similarity as follows:
 \begin{definition}[Graph Feature Similarity]
     We measure the graph feature similarity \(K\) through averaging the feature similarity of all its connected nodes pairs: $K  = \sum_{(u, v) \in \edgeSet} \tfrac{k(u,v)}{|\edgeSet|}$.
 \end{definition}

Unlike the homophily measures defined on node class labels which are non-negative~\cite{lim2021new}, the feature similarity $k(u, v) \in [-1, 1]$ can additionally be \emph{negative}, indicating negative correlations. We refer to the graph as \emph{positively correlated} if $K > 0$ and \emph{negatively correlated} if $K < 0$.

\section{Homophilic \& Heterophilic Link Prediction}
\label{sec:homo-hete-link-pred}

While existing works on node classifications usually define homophilic or heterophilic graphs by whether the majority of connected nodes share the same class labels~\cite{lim2021new,platonov2024characterizing}, we argue that the homophilic and heterophilic link prediction tasks should be defined based on how the distributions of feature similarity scores between connected and unconnected nodes are separated, as these definitions capture the fundamental differences on how link prediction scores are correlated with feature similarity scores;
\jiong{Mention this to abstract and intro}
within the category of homophilic or heterophilic tasks, we further show that the variation of the positive feature similarity scores affects the rate of change for the link prediction scores.
We present definitions with intuitive examples in \S\ref{sec:link-prediction-cat} and theoretical analysis in \S\ref{sec:homo-hete-opt}.

\subsection{Categorizing Link Prediction on Distributions of Feature Similarity}
\label{sec:link-prediction-cat}

We begin our discussion by considering the distributions of feature similarity scores for a random positive (edge) and negative (non-edge) node pair in the graph: consider the set of feature similarity scores for positive (edge) node pairs in the graph as $\mathcal{K}_{pos}$, and negative (non-edge) node pairs as $\mathcal{K}_{neg}$. We first formalize different categories of link prediction tasks, which are defined by how the distributions of $\mathcal{K}_{pos}$ and $\mathcal{K}_{neg}$ are (approximately) separated:

\begin{figure}[t!]
    \centering
    \begin{subfigure}{0.23\textwidth}
        \centering
        \includegraphics[keepaspectratio, width=\textwidth,trim={0 0 0 0},clip]{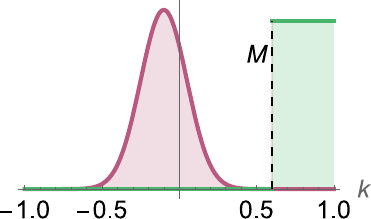}
        \caption{Homophilic, larger $M$}
        \label{fig:homo-narrow-vis}
    \end{subfigure}
    ~
    \begin{subfigure}{0.23\textwidth}
        \centering
        \includegraphics[keepaspectratio, width=\textwidth,trim={0 0 0 0},clip]{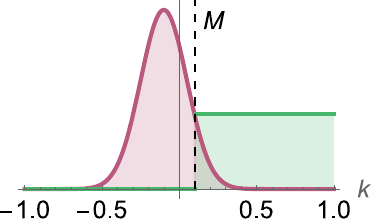}
        \caption{Homophilic, lower $M$}
    \end{subfigure}
    ~
    \begin{subfigure}{0.23\textwidth}
        \centering
        \includegraphics[keepaspectratio, width=\textwidth,trim={0 0 0 0},clip]{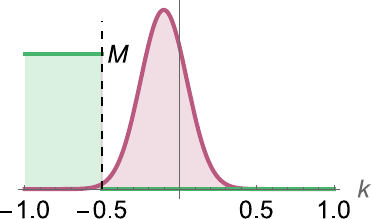}
        \caption{Heterophilic}
        \label{fig:hete-vis}
    \end{subfigure}
    ~
    \begin{subfigure}{0.23\textwidth}
        \centering
        \includegraphics[keepaspectratio, width=\textwidth,trim={0 0 0 0},clip]{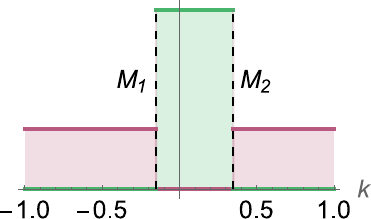}
        \caption{Gated}
        \label{fig:gated-vis}
    \end{subfigure}
    \caption{Categorizing link prediction tasks based on the distribution of feature similarity scores of positive node pairs (i.e., edges -- colored in green) and negative node pairs (non-edges -- colored in red): two distributions whose density is visualized in the plots are (approximately) separated by the threshold(s) $M$. Homophilic and heterophilic link prediction differs in whether the positive similarity scores fall into the larger or smaller side of the threshold $M$, while the magnitude of $M$ indicates the variance of positive similarity. Gated link prediction is a more complex case where the distribution of positive and negative similarity scores cannot be separated by a single threshold.}
    \label{fig:homo-hete-gated-vis}
\end{figure}

\begin{definition}[Homophilic Link Prediction] The task is \emph{homophilic} if $M \in \R$ exists such that for most samples $\tilde{\mathcal{K}}_{pos} \subset {\mathcal{K}}_{pos}$ and $\tilde{\mathcal{K}}_{neg} \subset {\mathcal{K}}_{neg}$ it satisfies $\inf(\tilde{\mathcal{K}}_{pos}) \geq M > \sup(\tilde{\mathcal{K}}_{neg})$. 
\end{definition}

Prior works have mostly focused on the homophilic category for the link prediction problem while overlooking other possibilities. For other cases where the homophilic conditions are not satisfied, we refer to them as \emph{non-homophilic} link prediction problems in general. In the definition below, we formalize an easy type of non-homophilic link prediction problem: 

\begin{definition}[Heterophilic Link Prediction] The task is \emph{heterophilic} if $M \in \R$ exists such that for most samples $\tilde{\mathcal{K}}_{pos} \subset {\mathcal{K}}_{pos}$ and $\tilde{\mathcal{K}}_{neg} \subset {\mathcal{K}}_{neg}$ it satisfies $\sup(\tilde{\mathcal{K}}_{pos}) \leq M < \inf(\tilde{\mathcal{K}}_{neg})$. 
\end{definition}

We give intuitive examples of homophilic and heterophilic link prediction tasks in Fig.~\ref{fig:homo-hete-gated-vis}: the key difference between homophilic and heterophilic tasks are whether $\mathcal{K}_{pos}$ is predominantly distributed on the larger side than $\mathcal{K}_{neg}$ (Fig.~\ref{fig:homo-narrow-vis} vs. \ref{fig:hete-vis}). The categorizations of homophilic/heterophilic link prediction tasks should not be confused with the magnitude of $M$ that indicates the variance of positive similarity scores: while its magnitude does not determine the type of the link prediction problem, our analysis in the next section does show that it affects the rate of change for the link prediction scores. 

However, beyond the heterophilic setting defined above, there are other non-homophilic settings with even more complexity, where the distribution of $\mathcal{K}_{pos}$ and $\mathcal{K}_{neg}$ cannot be separated by a single threshold $M$. 
Most of these cases are too complex to be formalized and studied theoretically, but we formalize one of them below (with visualization in Fig.~\ref{fig:gated-vis}) which we empirically experimented later:

\begin{definition}[Gated Link Prediction] The task is \emph{gated} if it is neither homophilic nor heterophilic, but $M_1, M_2 \in \R$ exist such that $M_2 \geq \sup(\tilde{\mathcal{K}}_{pos}) \geq \inf(\tilde{\mathcal{K}}_{pos}) \geq M_1$. 
    
\end{definition}

\subsection{Differences between Homophilic \& Heterophilic Link Prediction}
\label{sec:homo-hete-opt}

In \S\ref{sec:link-prediction-cat}, we have situated our discussions of link prediction tasks based on how the positive and negative samples are separated in the feature similarity score space. In this section, we reveal on a stylized learning setup the fundamental differences in optimizations for homophilic and heterophilic link prediction tasks. 

\begin{wrapfigure}{r}{0.4\textwidth}
    \centering
    \vspace{-0.4cm}
    \includegraphics[width=0.38\textwidth, trim={0 0.2cm 0 0cm}, clip]{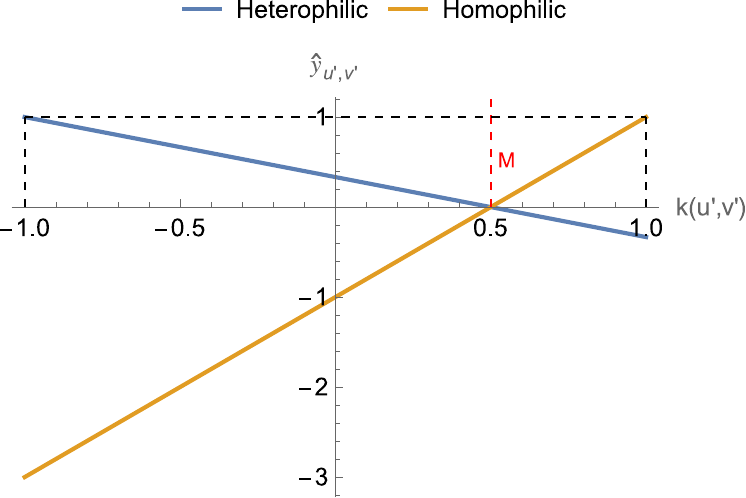}
    \caption{Link prediction scores $\hat{y}_{u'v'}$ for decoders optimized respectively under homophilic and heterophilic setups in Thm.~\ref{thm:hete-homo-sim-scores}, assuming $M=0.5$.}
    \vspace{-0.5cm}
    \label{fig:link-prob-comp}
\end{wrapfigure}
\paragraph{Theoretical Assumptions} \textit{Assuming a training graph with node $v \in \vertexSet$ whose feature vectors are 2-dimensional unit vectors, which can be represented as $\mathbf{x}_v = (\cos \theta_v, \sin \theta_v)$. We consider a DistMult decoder for predicting the link probability for candidate node pair $u, v \in \vertexSet$ with feature vectors $\mathbf{x}_u, \mathbf{x}_v$. Specifically, the link probability is calculated as $\hat{y}_{u,v} = (\mathbf{x}_u \otimes \mathbf{x}_v)^\T \mathbf{w} + b$, where $\mathbf{w}$ and $b$ are learnable parameters for the decoder, with training loss function $\mathcal{L}=y\cdot\mathrm{ReLU}(-\hat{y})+(1-y)\cdot\mathrm{ReLU}(\hat{y})$ such that $\hat{y} \geq 0$ for all edges (positive samples) and $\hat{y} < 0$ otherwise.
Furthermore, we assume that the $\mathcal{K}_{pos}$ and $\mathcal{K}_{neg}$ are \emph{ideally separable} by a threshold $M\in[0,1]$ in the feature similarity score space such the homophilic or heterophilic conditions hold for \emph{all} samples.}

With the above assumptions, we now show that (1) the predicted link probability and feature similarity scores are positively correlated for homophilic tasks, while negatively correlated for heterophilic tasks; (2) the change rate for the predicted link probability with respect to the feature similarity is determined by the magnitude of the threshold $M$ that separates the positive and negative samples.

\begin{theorem} 
    \label{thm:hete-homo-sim-scores}
    Following the above assumptions, consider two DistMult decoders that are fully optimized for homophilic and heterophilic link prediction problems respectively. Give an arbitrary node pair $(u', v')$ with node features $\mathbf{x}_{u'} = (\cos \theta_{u'}, \sin \theta_{u'})$ and $\mathbf{x}_{v'} = (\cos \theta_{v'}, \sin \theta_{v'})$ and pairwise feature similarity $k(u', v')$, the following holds for the predicted link probability $\hat{y}_{u'v'}$:
    \begin{itemize}[leftmargin=*]
        \item For homophilic problem where $1 \geq \inf(\mathcal{K}_{pos}) = M > \sup(\mathcal{K}_{neg})$, when bounding $\hat{y}_{u,v}=1$ if $k(u, v)=1$ during training, $\hat{y}_{u'v'}$ increases with $k(u', v')$ at a linear rate of $\tfrac{1}{(1-M)}$;
        \item For heterophilic problem where $-1 \leq \sup(\mathcal{K}_{pos}) = M < \inf({\mathcal{K}}_{neg})$, when bounding $\hat{y}_{u,v}=1$ if $k(u, v)=-1$ during training, $\hat{y}_{u'v'}$ decreases with $k(u', v')$ at a linear rate of $\tfrac{1}{(1-M)}$.
    \end{itemize}
\end{theorem}

We give proof in App. \S\ref{app:sec:proof-hete-homo-sim-scores} and visualize in Fig.~\ref{fig:link-prob-comp} how the predicted link probability $\hat{y}_{u'v'}$ changes under homophilic and heterophilic settings. Though the above results are derived under simplified assumptions, it clearly highlights the different optimizations needed for homophilic and heterophilic link prediction tasks that have not been studied in prior literature.
In \S\ref{sec:exp}, we observe that these differences go beyond our theoretical assumptions and affect the performance of all GNN encoders and link prediction decoders on datasets with more complexity, which warrant the our study of effective encoder and decoder choices for non-homophilic link prediction in the next section.

\section{Encoder \& Decoder Choices for Link Prediction Beyond Homophily}
\label{sec:encoder-decoder-choices}

In~\S\ref{sec:homo-hete-link-pred}, we gave formal definitions of homophilic and heterophilic link prediction tasks and highlight their differences in model optimizations. As non-homophilic settings are largely overlooked in prior literature, we aim to verify whether existing GNN message passing designs for node features remain effective beyond homophily. 
We follow the encoder-decoder perspective in \cite{hamilton2017representation} and discuss designs for both GNN encoder and link prediction decoder that adapt to non-homophilic settings. 

\subsection{Decoder Choice for Heterophilic \& Gated Link Prediction}
\label{sec:decoder-choices}

For homogeneous graphs, popular decoder choices for deriving link probability from node representations are either a simple dot product (DOT) operation or more complex multi-layer perceptron (MLP). 
While MLP decoder has a stronger representation power due to its non-linearity, inner product decoder is more preferred in large-scale applications due to its fast inference speed: it is well established that maximum inner product search (MIPS) can be approximated with sublinear complexity using packages such as Faiss~\cite{douze2024faiss}. A prior work~\cite{wang2022flashlight} has benchmarked the performance of different link prediction decoders on several OGB datasets and proposed a sublinear approximation of MLP decoder during inference time; however, no study has been conducted on the performance of decoders for non-homophilic link prediction tasks across the negative to positive similarity spectrum.

Our takeaways for effective decoder choices for non-homophilic link prediction tasks are as follows: 
(1) for non-homophilic (e.g., gated) tasks, only non-linear decoders such as MLP are suitable; (2) for heterophilic tasks, a linear decoder with learnable weights (e.g., DistMult~\cite{yang2014embedding}) can be used in lieu of MLP to achieve better scalability while maintaining comparable performance; 
(3) dot product decoder is only suitable for homophilic link prediction tasks.

Theoretically, we formalize our first takeaway with the below theorem, which shows the limitations of using linear decoders (such as DOT and DistMult) in non-homophilic link prediction tasks:

\begin{theorem}
    \label{thm:linear-decoder-limitation}
    No parameter exists for a single linear decoder that perfectly separates link probability for edges and non-edges for gated link prediction. 
\end{theorem}

We present the proof in Appendix \S\ref{app:sec:proof-linear-decoder-limitation}. For linear models, while both DistMult and DOT product decoders share the same time complexity during inference, we observe empirically in \S\ref{sec:exp} that DistMult outperforms DOT decoder by up to 55\% on non-homophilic link prediction tasks. Intuitively, the learnable weights in DistMult decoder allow the model to capture the negative correlation between connected node features and improve its effectiveness for heterophilic tasks.

\subsection{Improving GNN Representation Power with Heterophily-adjusted Designs}
\label{sec:encoder-choices}

We now consider the impact of GNN architectures on non-heterophilic link prediction performance. 
In particular, we examine whether the effective designs for node classification beyond \emph{class} homophily can be transferred to link prediction tasks beyond \emph{feature} homophily. A design that significantly improves classification performance under low class homophily is the separation of ego- and neighbor-embeddings in GNN message passing, which has consistently shown to improve classification performance across multiple studies~\cite{zhu2020beyond,platonov2022critical}. 
As real-world graphs usually follow a power-law degree distribution and exhibit large variation in node degrees, prior work has used the robustness of GNN models to degree shift as a proxy to measure the generalization ability of GNN models for heterophilic node classification~\cite{zhu2020beyond}.
We follow a similar approach in the theorem below and show that 
a GNN model that embeds ego- and neighbor-features together is less capable of generalizing under heterophilic settings than a graph-agnostic model. 
We give the proof in Appendix \S\ref{app:sec:proof-ego-neighbor-separation}. 

\begin{theorem}
    \label{thm:ego-neighbor-separation}
    Consider the same DistMult decoder and loss function $\mathcal{L}$ as the assumptions in \S\ref{sec:homo-hete-opt}, but trained on a heterophilic graph where (1) feature vectors for all nodes can be either $\mathbf{x}_1 = (\cos \theta_1, \sin \theta_1 )$ or $\mathbf{x}_2 = (\cos \theta_2, \sin \theta_2)$, and (2) nodes $u$ and $v$ are connected if and only if $\mathbf{x}_u \neq \mathbf{x}_v$.
    Assume two DistMult decoders are trained, one baseline with node features $\mathbf{x}_u$, and the other with GNN representations $\mathbf{r}_u$ instead of node features $\mathbf{x}_u$, where $\mathbf{r}_u$ is obtained with a linear GNN model $\mathbf{r}_u = \frac{1}{| \neighNoSelfLoop(u)| + 1} \mathbf{x}_u + \frac{1}{| \neighNoSelfLoop(u)| + 1} \sum_{l \in \neighNoSelfLoop(u)}\mathbf{x}_l$ that considers self-loops in its message passing process. We further assume a degree shift between training and test sets, where all training nodes have degree $d$ while the test nodes have degree $d'$. Then for any $d'>0$ when $d=0$, or $1 \leq d' < d$ when $d \geq 2$, the DistMult decoder optimized on GNN representations $\mathbf{r}_u$ reduce the separation distance between edges and non-edges for the test nodes compared to the baseline optimized with node features $\mathbf{x}_u$. 
\end{theorem}

\section{Related Work}
\label{sec:related}

\textbf{Graph Neural Networks for Link Prediction.} Traditional algorithms for link prediction are primarily \textit{heuristic-based}, which have strong assumptions on the link generation process. These approaches compute the similarity scores between two nodes based on certain structures or properties~\cite{adamic2003friends, barabasi1999emergence, brin1998anatomy, zhang2022graph}. Later, various \textit{representation-based} algorithms for link prediction have been proposed, which primarily aim at learning low-dimensional node embeddings that are used to predict the likelihood of link existence between certain node pairs and usually involve the use of GNNs~\cite{kipf2016semi,hamilton2017inductive, velickovic2018graph}. Compared with the \textit{heuristic-based} algorithms, \textit{representation-based} algorithms do not require strong assumptions and perform learning of the graph structure and node features in a unified way. A representative method is the Variational Graph Autoencoder~\cite{kipf2016variational}, which uses GCN as the encoder for learning node representations and inner product as the decoder for pairwise link existence predictions. More recently, the state-of-the-art methods for link prediction are built on top of the \textit{representation-based} algorithms and augment them with additional \textit{pairwise} information. For instance, \textit{subgraph-based} approaches perform link prediction between two nodes by first extracting their enclosing subgraphs and then followed by applying the standard \textit{representation-based} algorithms on the extracted subgraphs~\cite{zhang2018link,chamberlain2022graph-ct,zhu2021neural}. Concurrently, other works have been proposed to augment GNN learning with common neighbor information~\cite{wang2023neural,yun2022neo-gnns-cr,wang2023neural}. Despite the tremendous success, prior link prediction works mainly hold the \emph{homophilic} assumption, where node pairs with similar features or neighbors are more likely to link together\jiong{revisit}. In contrast, this work characterizes a notion of heterophily in link prediction and explore how popular approaches perform under our characterization. Related work also examines the influence of data for link prediction from a joint perspective of graph structure and feature proximity~\cite{mao2023revisiting,lee2024netinfof}. On the contrary, our work considers the role of features more carefully and exclusively from the perspective of heterophily.

\textbf{Graph Neural Networks Addressing Heterophily.} There has been a rich literature on graph neural networks facing heterophily, but most of them are in a node classification setting~\cite{abu2019mixhop,zhu2020graph, zhu2020beyond,pei2020geom,he2021bernnet,luan2022revisiting}\jiong{add more citations}. Very few works pay attention to the problem of link prediction under heterophily. Among them, \cite{zhou2022link} proposes to disentangle the node representations from latent dissimilar factors. \cite{di2024link} extends the physics-inspired GRAFF~\cite{di2022understanding}, originally designed for handling heterophilic node classification tasks, to heterophilic link prediction tasks. These works use the same homophily measure as typically used in node classification, while our work emphasizes the influence of features in link prediction and more systematically benchmarks model performance against different homophily levels.    
\section{Empirical Analysis}
\label{sec:exp}

\newcommand{\githubrepo}{\url{https://github.com/tensor-gales/HeteLinkPred}}

We aim to understand through empirical analysis (1) what are the performance trends of link prediction methods under different link prediction tasks in the full spectrum of negative to positive feature similarity, and (2) how different encoder and decoder designs adapt to non-homophilic link prediction tasks, including variations of feature similarity and node degrees within the same graph. We first introduce the link prediction methods that we consider in our experiments, and then present the results on synthetic and real-world datasets. More details about setups and results are available in App.~\ref{app:sec:exp-details}.

\paragraph{Link Prediction Methods} 
As in the previous sections, we follow an encoder-decoder framework and consider different combinations of both components. For \textit{decoders}, we consider the options mentioned in \S\ref{sec:decoder-choices}: (1) \textbf{Dot Product (DOT)}, (2) \textbf{Multi-Layer Perceptron (MLP)}, and (3) \textbf{DistMult}~\cite{yang2014embedding}. 

For \textit{encoders}, to validate how the design of ego- and neighbor-embedding separation affects link prediction performance, we follow the analysis of \cite{zhu2020beyond} and consider two classic GNN methods: (1) \textbf{GraphSAGE}~\cite{hamilton2017inductive} which features this design, and (2) \textbf{Graph Convolutional Network (GCN)}~\cite{kipf2016semi} which does not. We couple these encoders with the decoders above to form six GNN4LP models.

Furthermore, to understand how the aforementioned design in message passing of features affect performance for GNN4LP models that also leverage pairwise structural information, we consider a SOTA method~\cite{li2023evaluating-lb}---BUDDY~\cite{chamberlain2022graph-ct}---which augments GNN encoders with subgraph sketching. Specifically, we consider the following variants of BUDDY: (3) \textbf{BUDDY-GCN}, which uses GCN for feature aggregation, (4) \textbf{BUDDY-SIGN}, which uses the SIGN aggregation~\cite{rossi2021knowledge} that separates ego- and neighbor-embeddings in message passing, and (5) a feature-agnostic \textbf{NoGNN} baseline.
We note that BUDDY has a MLP decoder built in its design.

Finally, we also consider the following link prediction heuristics tested in \cite{chamberlain2022graph-ct}: \textbf{Common Neigbhors (CN)}, \textbf{Resource Allocation (RA)}, \textbf{Adamic-Adar (AA)}, and \textbf{Personalized PageRank (PPR)}.

\subsection{Experiments on Synthetic Graphs}

We generate synthetic graphs that resemble different types of link prediction tasks (i.e., homophilic, heterophilic, and gated) by varying the feature similarity between connected nodes. 
These graphs provide controlled environments that allow us to focus on the effects of feature similarity on link prediction performance without mingling them with other data factors such as structural proximity. We give the details of the synthetic graph generation process and the experiment setup in App. \S\ref{app:sec:exp-details}.

\begin{figure}[t!]
    \centering
    \begin{subfigure}{0.32\textwidth}
        \centering
        \includegraphics[keepaspectratio, width=\textwidth,trim={0 0 0 0},clip]{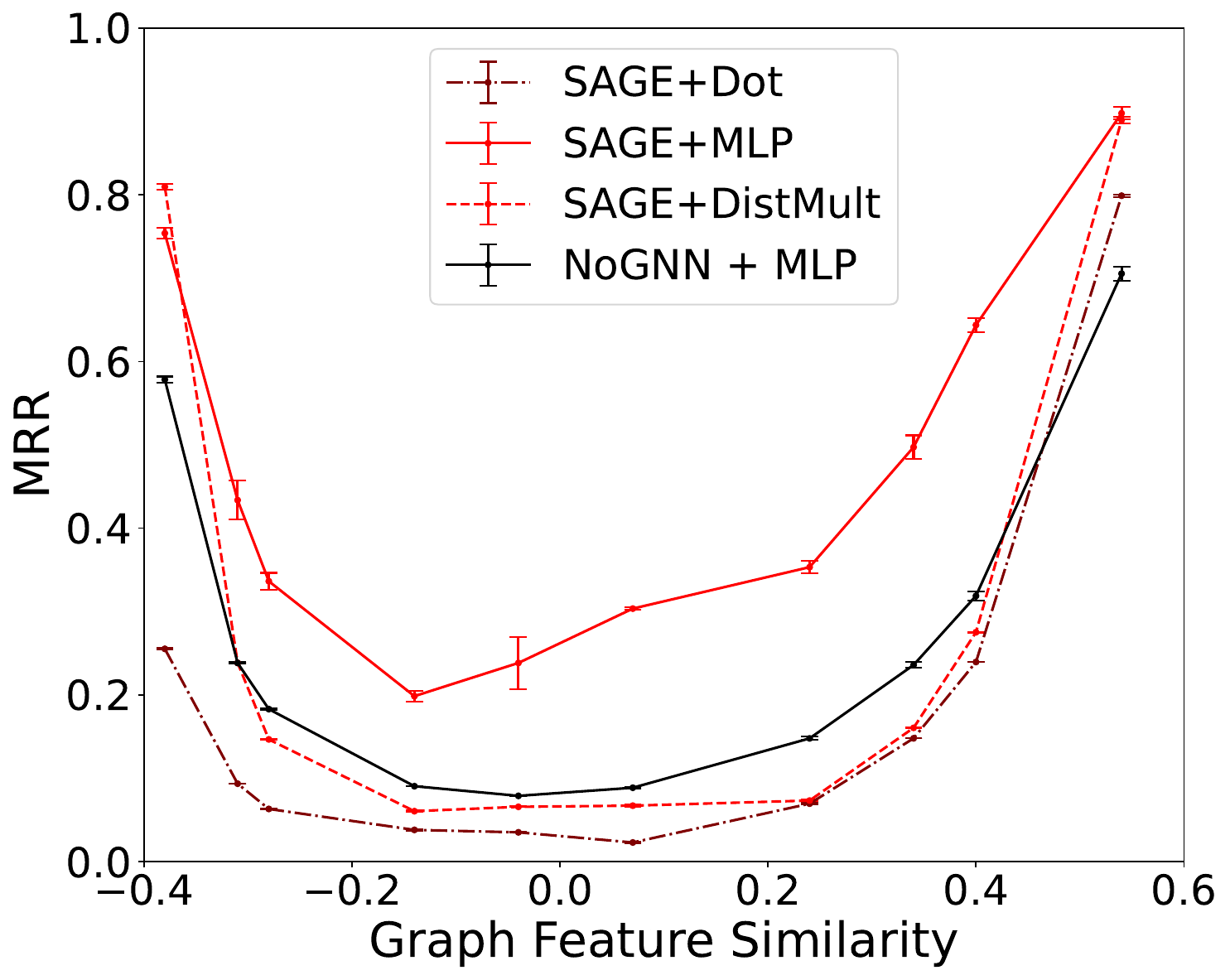}
        \caption{Comparison of decoder choices with SAGE encoder.}
        \label{fig:synthetic-decoder-fixed-sage}
    \end{subfigure}
    ~
    \begin{subfigure}{0.32\textwidth}
        \centering
        \includegraphics[keepaspectratio, width=\textwidth,trim={0 0 0 0},clip]{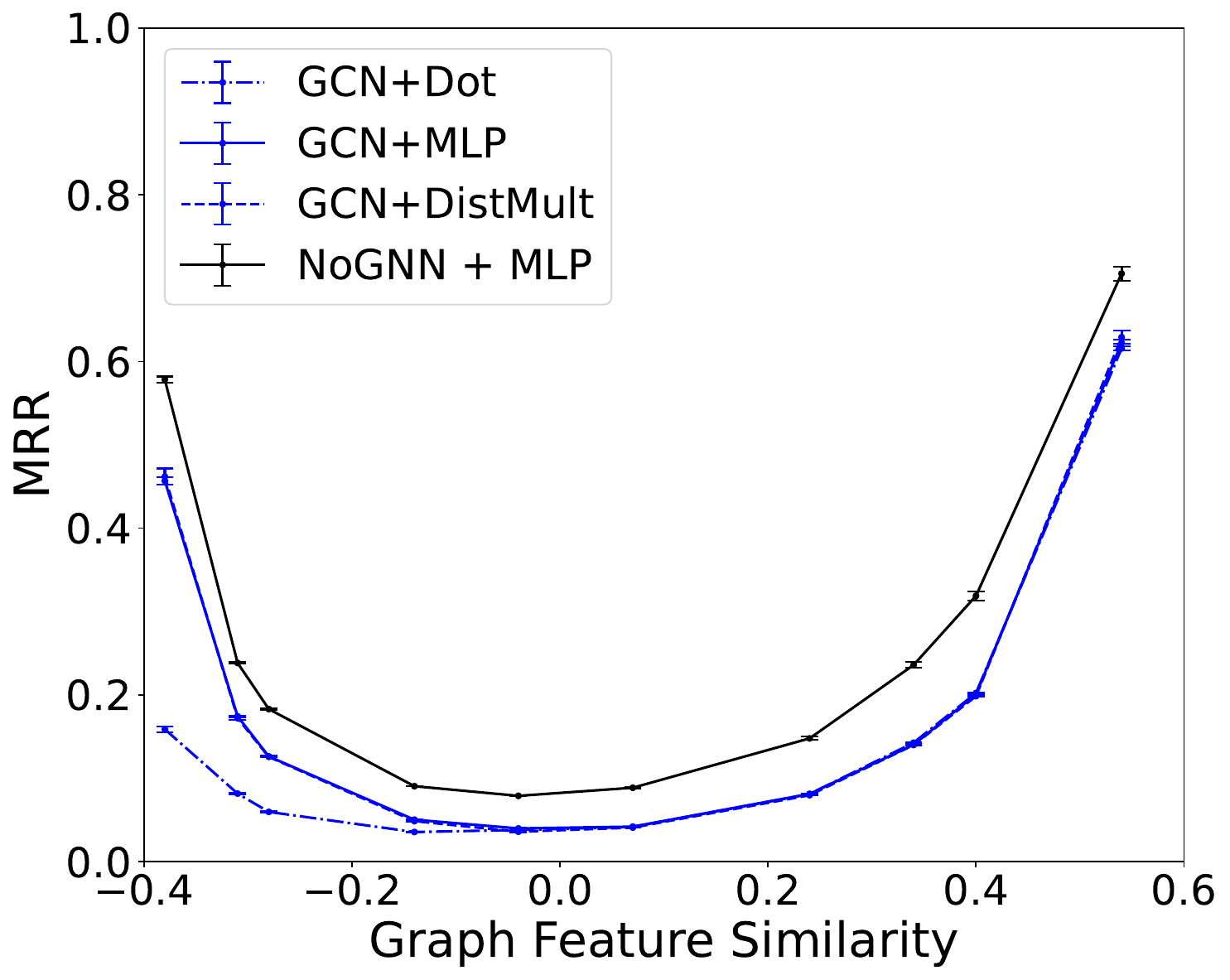}
        \caption{Comparison of decoder choices with GCN encoder.}
        \label{fig:synthetic-decoder-fixed-gcn}
    \end{subfigure}
    ~
    \begin{subfigure}{0.32\textwidth}
        \centering
        \includegraphics[keepaspectratio, width=\textwidth,trim={0 0 0 0},clip]{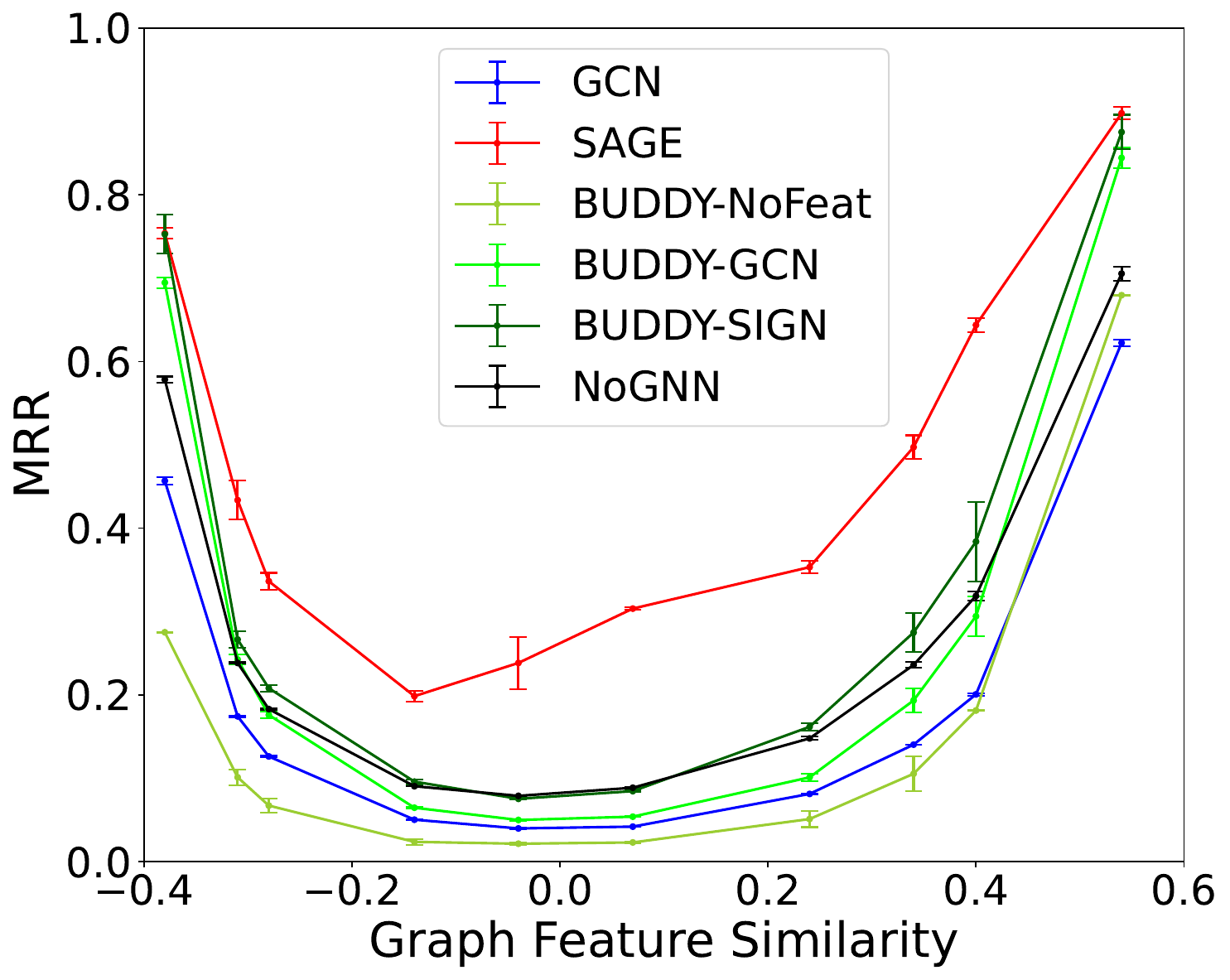}
        \caption{Comparison of different encoder choices with MLP decoder.}
        \label{fig:synthetic-encoder-fixed-mlp}
    \end{subfigure}
    \caption{Comparing link prediction methods on synthetic graphs with varying levels of feature similarity: (\subref{fig:synthetic-decoder-fixed-sage}) and (\subref{fig:synthetic-decoder-fixed-gcn}) focus on decoders, while (\subref{fig:synthetic-encoder-fixed-mlp}) focuses on encoders. 
    We include MLP decoder without GNN as a graph agnostic baseline in all plots. 
    Numerical results are reported in Table~\ref{tab:synthetic-results}. 
    }
    \vspace{-0.2cm}
    \label{fig:synthetic-results}
\end{figure}

\paragraph{Performance Trend Across Similarity Specturm}  
We report visualize their performance trends per method in Fig.~\ref{fig:synthetic-results} and defer the numerical results to Table~\ref{tab:synthetic-results}. 
We observe that the performance of all feature-consuming methods is significantly affected by the level of feature similarity: most methods reach their best performance at the positive extreme (homophilic tasks) and the second best at the negative extreme (heterophilic tasks); between the two extremes (gated tasks), the performance of all methods drops significantly as the feature similarity score approaches 0, which creates a U-shaped performance trend across the feature similarity spectrum.
It is worth noting that graph-agnostic MLP decoder without GNN (NoGNN) also exhibits the U-shaped performance trend, which suggests the challenges of leveraging node features effectively in these settings for non-GNN methods as well. Intuitively, as the similarity scores for a random pair of nodes follow a normal distribution centered around 0, the performance drop when average feature similarity scores approach 0 can be contributed to the reduced distinguishability between the similarity scores of edges and non-edges in the graph. 
For feature agnostic heuristics such as Common Neighbors and Personalized PageRank, they show mostly unproductive performance except at the positive extreme. This suggests that graphs formed by positive feature correlations are also likely to show strong structural proximity, which is beneficial for the heuristic-based methods for the homophilic link prediction task.

\paragraph{Decoder Choices: MLP and DistMult over DOT}
We further validate our takeaways in \S\ref{sec:encoder-decoder-choices} regarding how different decoder choices adapt to non-homophilic link prediction settings: In Fig.~\ref{fig:synthetic-decoder-fixed-sage}-\subref{fig:synthetic-decoder-fixed-gcn}, we compare the performance of different decoders with fixed SAGE and GCN encoders, respectively. 
With both SAGE and GCN encoders, we observe that DOT decoder performs the worst among all choices across all feature similarity levels: it is outperformed by MLP decoder with a margin of 50\% in the negative extreme and 10\% in the positive extreme under SAGE encoder.
DistMult decoder performs significantly better than the DOT decoder, especially in the region of negative feature similarity: at the negative extreme, DistMult decoder outperforms DOT decoder by 55\% and even outperforms MLP decoder by 5.6\%. Empirically, we observe that the optimization process of DistMult is more stable than MLP when using a SAGE encoder in the negative similarity region, allowing it to reach optimal performance without suffering from instabilities at the negative extreme. However, the performance of DistMult coupled with SAGE decoder is significantly lower than MLP with up to 37\% gap between the negative and positive extremes, where the link prediction tasks are gated instead of being homophilic or heterophilic. This validates our theoretical analysis in \S\ref{sec:decoder-choices} that the linear decoders like DOT and DistMult are not suitable for the settings where non-linear separation between similarity scores of edges and non-edges are required. 
With GCN encoder, the performance of DistMult decoder is mostly on par with MLP decoder across the specturm, which shows that the performance bottleneck is on the encoder side rather than the decoder side. 
Overall, MLP decoder is the most robust choice across different feature similarity levels and link prediction tasks, yielding the best performance in all but one cases when coupled with SAGE encoder, with DistMult being a more scalable alternative for homophilic and heterophilic link prediction tasks.

\textbf{Encoder Choices: Importance of Ego- and Neighbor-Embedding Separation.}
In Fig.~\ref{fig:synthetic-encoder-fixed-mlp}, we compare the performance of different encoder choices with fixed MLP decoder, which is the best-performing decoder option for nearly all cases. In addition to GCN and SAGE encoders that rely only on node features, we also include variants of BUDDY~\cite{chamberlain2022graph-ct} that leverage structural proximity and (optionally) node features.
Comparing between SAGE and GCN encoders, we observe that SAGE consistently outperforms GCN across all feature similarity levels by up to 44\%. For BUDDY variants with SIGN and GCN feature encoders, we also observe consistently better performance with SIGN encoder across all feature similarity levels with up to 9.0\% gain. Both comparisons suggest the importance of adapting ego- and neighbor-embedding separation in GNN encoder design for link prediction: as discussed in \S\ref{sec:encoder-choices}, this design allows GNN encoders to learn representations that are more robust to variations of node degrees and feature similarity levels in the graph instead of overfitting to specific degree or similarity buckets\jiong{revisit \S\ref{sec:encoder-choices} to align the story}, which we also observe in the real-world datasets. 

\paragraph{Importance of Node Features vs. Structural Proximity} 
Comparing the best feature-only method (SAGE+MLP) with the best-performing BUDDY variant (BUDDY-SIGN) which additionally captures structural proximity captured through subgraph sketching, we find that the performance of BUDDY-SIGN is consistently lower than SAGE+MLP across all feature similarity levels, especially between the negative and positive extremes where the gap is up to 26\%. This shows that the node features can be more informative than structural proximity for link prediction tasks when graph connections are predominantly driven by feature similarity. While the underlying factors that drive connections in the real-world graphs are more complex and can involve both feature similarity and structural proximity, these results highlight the importance for understanding the effective use of node features in link prediction tasks and avoid over-reliance on structural proximity especially for non-homophilic cases.

\subsection{Experiments on Real-world Graphs}

Next we compare performance of different link prediction methods on real-world graphs of varying sizes and feature similarities. Unlike our synthetic graphs where the feature similarity of connected nodes is controlled, real-world graphs tend to have a much larger variation in feature similarity across edges, and the performance discrepancy of different methods on edges with different feature similarity are often overlooked in the literature. Our analysis aims to not only compare the overall performance of the methods, but also to understand their robustness to similarity variations in the graph.

\paragraph{Experiment Setups} 
We consider three real-world datasets: ogbl-collab~\cite{hu2020ogb}, ogbl-citation2~\cite{hu2020ogb}, and e-comm~\cite{zhu2024pitfalls}. The details of these datasets and our experiment setup are presented in App.~\S\ref{app:sec:exp-details}. We report the dataset statistics in Table~\ref{tab:real-world-results}. Furthermore, we show the feature similarity distributions of edges vs. random node pairs in App. Fig.~\ref{fig:real-graph-sim-hist}: while link predictions on ogbl-collab and ogbl-citation2 can be approximately seen as homophilic tasks, link prediction on e-comm is non-homophilic.

In addition to the overall performance, we also examine the performance of different encoder and decoder choices on groups of edges bucketized by degrees of connected nodes and feature similarity scores, and calculate the average link prediction performance of each method on the edges within each bucket; we give more details in App.~\S\ref{app:sec:exp-details}. To compare how different encoder and decoder choices adapt to these local property variations, we conduct the analysis by fixing either the encoder or decoder and visualizing the performance gain and loss per bucket by varying the other choice.

\begin{table}[t]
    \small
    \begin{minipage}{0.52\textwidth}
        \centering
        \caption{Results on real-world graphs. "*" denotes results quoted from \cite{chamberlain2022graph-ct}.}
        \label{tab:real-world-results}
        \begin{tabular}{rrrrcccccccc}
            \toprule 
            \textbf{Dataset} & \textbf{ogbl-collab} & \textbf{ogbl-citat2} & \textbf{e-comm} \\
            \textbf{\#Nodes} & 235,868 & 2,927,963 & 346,439\\
            \textbf{\#Edges} & 2,358,104 & 30,387,995 & 682,340\\
            \textbf{Feat. Sim} & $0.70{\scriptstyle \pm 0.23}$ & $0.40{\scriptstyle \pm 0.22}$ & $0.18{\scriptstyle \pm 0.63}$ \\
            \textbf{Metrics} & Hits@50 & MRR & MRR \\
            \midrule
            & \multicolumn{3}{c}{\textsc{Heuristics}} \\
            \cmidrule{2-4}
            \textbf{CN} & 56.44* & 51.47* & 0.20 \\
            \textbf{AA} & 64.35* & 51.89* & 0.20 \\
            \textbf{RA} & 64.00* & 51.98* & 0.20 \\
            \midrule
            & \multicolumn{3}{c}{\textsc{DOT Decoder}} \\
            \cmidrule{2-4}
            \textbf{GCN} & $10.64{\scriptstyle \pm 0.42}$ & $40.38{\scriptstyle \pm 1.52}$ & $33.83{\scriptstyle \pm 1.34}$ \\
            \textbf{SAGE} & $19.71{\scriptstyle \pm 0.52}$ & $71.39{\scriptstyle \pm 0.28}$ & $52.30{\scriptstyle \pm 5.11}$ \\
            \textbf{GPRGNN} & $14.28 {\scriptstyle \pm 2.52}$ & $55.82 {\scriptstyle \pm 3.82}$
            & $35.62 {\scriptstyle \pm 1.22}$ \\
            \midrule
            & \multicolumn{3}{c}{\textsc{DistMult Decoder}} \\
            \cmidrule{2-4}
            \textbf{GCN} & $25.62{\scriptstyle \pm 0.82}$ & $62.31{\scriptstyle \pm 0.68}$ & $53.09{\scriptstyle \pm 2.37}$ \\
            \textbf{SAGE} & $43.50{\scriptstyle \pm 1.13}$ & $82.26{\scriptstyle \pm 0.02}$ & $50.15{\scriptstyle \pm 6.56}$ \\
            \textbf{GPRGNN} & $28.42 {\scriptstyle \pm 4.22}$ & $72.88 {\scriptstyle \pm 1.89}$ & $45.55 {\scriptstyle \pm 3.18}$ \\
            \midrule
            & \multicolumn{3}{c}{\textsc{MLP Decoder}} \\
            \cmidrule{2-4}
            \textbf{NoGNN} & $6.07{\scriptstyle \pm 0.18}$ & $27.64{\scriptstyle \pm 0.21}$ & $24.65{\scriptstyle \pm 0.21}$ \\
            \textbf{GCN} & $30.17{\scriptstyle \pm 2.90}$ & $73.57{\scriptstyle \pm 0.35}$ & $54.82{\scriptstyle \pm 3.42}$ \\
            \textbf{SAGE} & $48.64{\scriptstyle \pm 0.39}$ & $83.67{\scriptstyle \pm 0.07}$ & $54.60{\scriptstyle \pm 0.09}$ \\
            \textbf{GPRGNN} & $39.52{\scriptstyle \pm 1.28}$ 
            & $84.55 {\scriptstyle \pm 3.55}$ & $52.77 {\scriptstyle \pm 3.64}$\\
            \midrule
            & \multicolumn{3}{c}{\textsc{BUDDY with MLP Decoder}} \\
            \cmidrule{2-4}
            \textbf{NoFeat} & $66.06{\scriptstyle \pm 0.22}$ & $83.36{\scriptstyle \pm 0.14}$ & $6.68{\scriptstyle \pm 0.00}$ \\
            \textbf{GCN} & $66.21{\scriptstyle \pm0.33}$ & $87.05{\scriptstyle \pm 0.04}$ & $13.13{\scriptstyle \pm 1.91}$ \\
            \textbf{SIGN} & $66.64{\scriptstyle \pm0.64}$ & $87.53{\scriptstyle \pm 0.12}$ & $10.95{\scriptstyle \pm 3.68}$ \\
            \bottomrule
        \end{tabular}
    \end{minipage}
    ~
    \begin{minipage}{0.47\textwidth}
        \begin{subfigure}[t]{0.48\textwidth}
            \centering
            \textbf{(a)-(c) ogbl-citat2}
            \includegraphics[width=\textwidth]{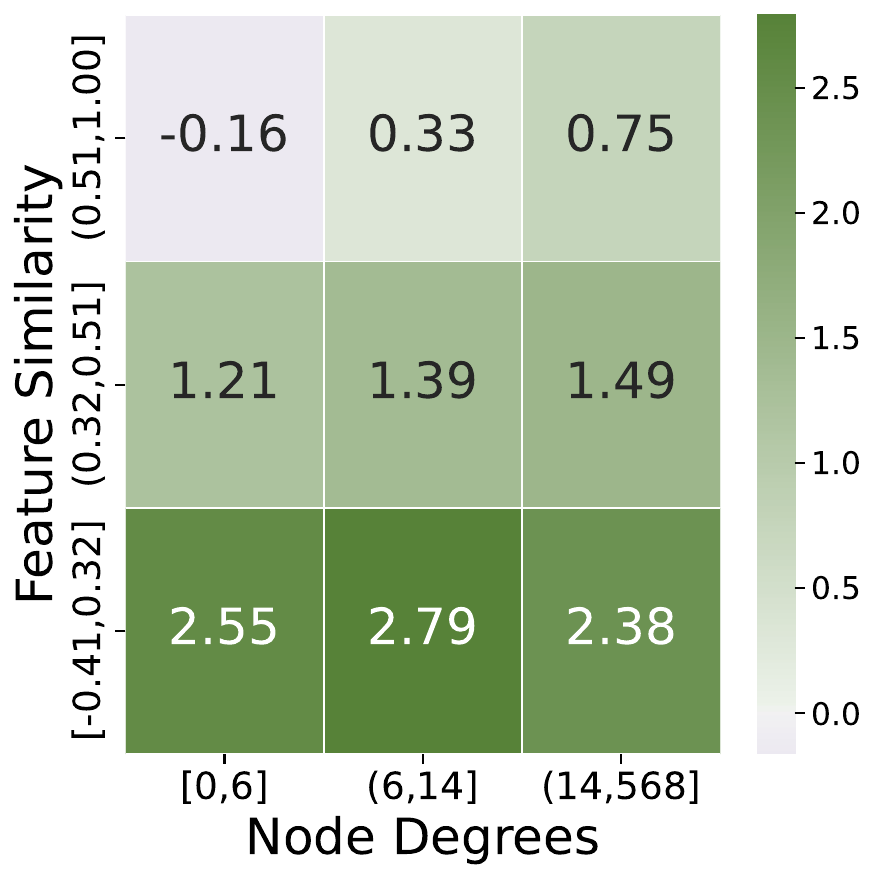}
            \caption{$\mathtt{mrr}_{\mathrm{MLP}} - \mathtt{mrr}_{\mathrm{DistM}}$ with SAGE encoder.}
            \label{fig:per-edge-main-citation2-mlp-distmult}


            \includegraphics[width=\textwidth]{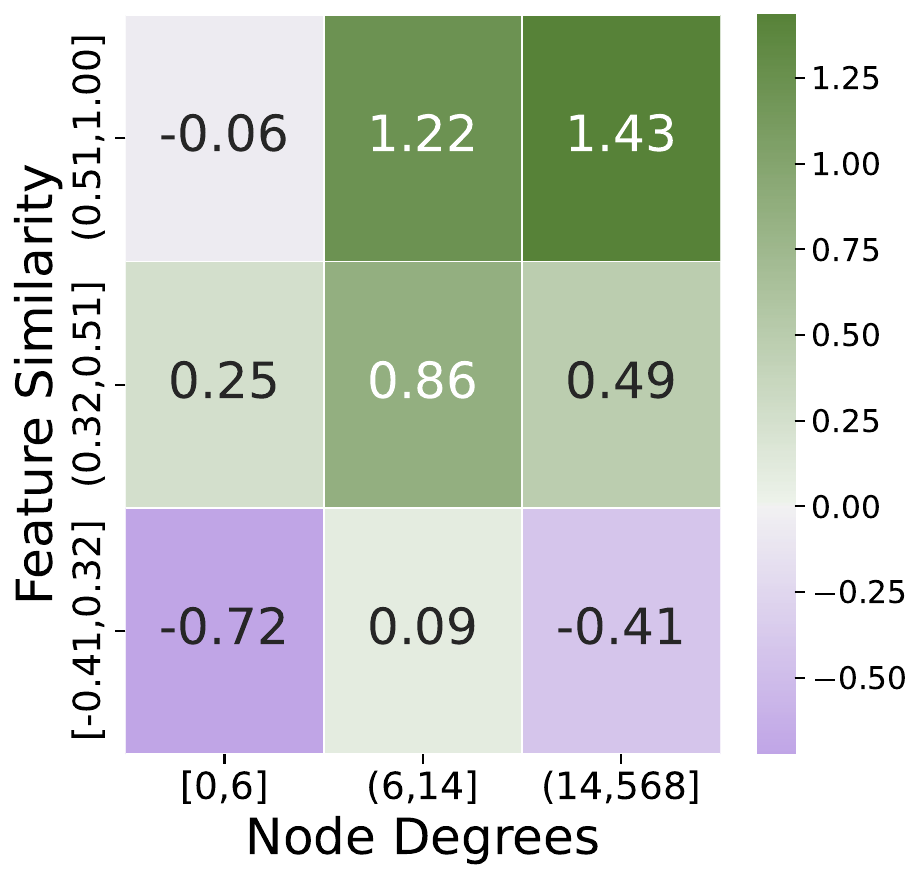}
            \caption{$\mathtt{mrr}_{\mathrm{SIGN}} - \mathtt{mrr}_{\mathrm{GCN}}$ with BUDDY \& MLP.}
            \label{fig:per-edge-main-citation2-sign-gcn}
        \end{subfigure}
        ~
        \begin{subfigure}[t]{0.46\textwidth}
            \centering
            \textbf{(d)-(f) e-comm}
            \includegraphics[width=\textwidth]{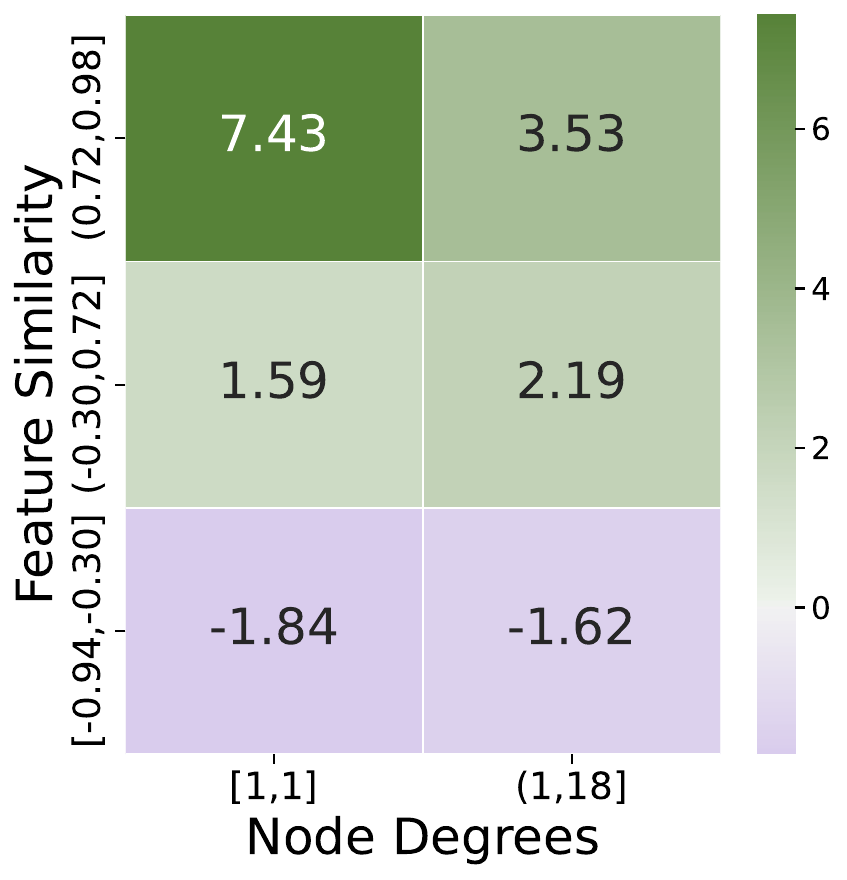}
            \caption{$\mathtt{mrr}_{\mathrm{MLP}} - \mathtt{mrr}_{\mathrm{DOT}}$ with SAGE encoder.}
            \label{fig:per-edge-main-ecomm-mlp-dot}

            \includegraphics[width=\textwidth]{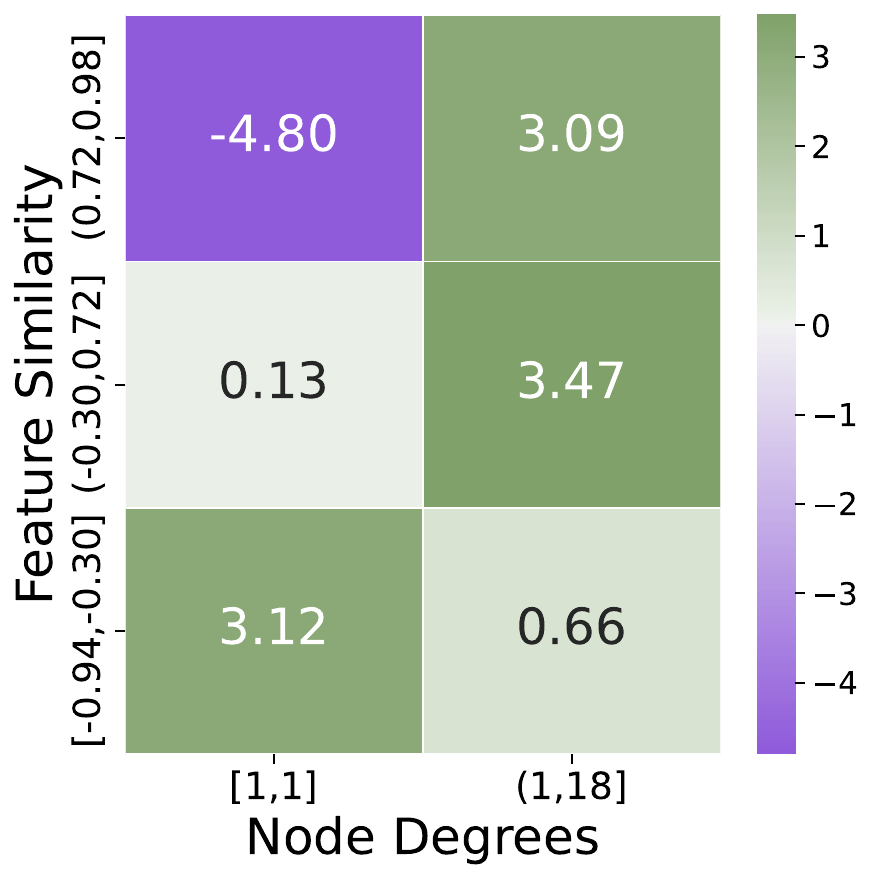}
            \caption{$\mathtt{mrr}_{\mathrm{SAGE}} - \mathtt{mrr}_{\mathrm{GCN}}$ with MLP decoder.}
            \label{fig:per-edge-main-ecomm-sage-gcn}

        \end{subfigure}

        \captionof{figure}{Pairwise comparison of encoder or decoder choices on test edges grouped by node degrees (x-axis) and feature similarity (y-axis): Green denotes MRR increases and purple denotes decreases. More plots in Fig.~\ref{fig:per-edge-analysis-count}-\ref{fig:per-edge-analysis-esci}.}
        \label{fig:per-edge-main}
    \end{minipage}
    \vspace{-0.5cm}
\end{table}

\textbf{Significance of Decoder Choices.}
We first compare the performance of different decoder choices while fixing the encoder. As we observed on the synthetic datasets, for both GCN and SAGE encoders, MLP has the highest overall performance, followed by DistMult and DOT. 
In terms of the robustness to local variations of node degrees and feature similarity scores, we observe in Fig.~\ref{fig:per-edge-main}\subref{fig:per-edge-main-ecomm-mlp-dot} that despite MLP only outperforms DOT by 2.3\% on the full test split, it outperforms DOT on the majority of the feature similarity range by up to 7.4\% as DOT overfits to the lowest feature similarity bucket; we also observe similar overfitting of DistMult to the highest feature similarity bucket in Fig.~\ref{fig:per-edge-main}\subref{fig:per-edge-main-citation2-mlp-distmult}, despite it being much more robust than DOT. Overall, MLP is the most robust choice for link prediction tasks on real-world graphs with varying feature similarity levels. 

\textbf{Significance of Encoder Choices.}
We finally compare different encoder choices while fixing the decoder. Matching our observations on synthetic datasets, we find that SAGE encoder outperforms the GCN encoder by significant margin under most datasets and decoder choices. On e-comm dataset, the performance of GCN on the full test split is on-par with SAGE under MLP decoder, but we observe in Fig.~\ref{fig:per-edge-main}\subref{fig:per-edge-main-ecomm-sage-gcn} that GCN largely overfits to the edges in the high feature similarity and low degree bucket, with SAGE outperforming GCN by up to 3.5\% in the remaining buckets. The similar overfitting is also observed on SIGN vs. GCN with BUDDY and MLP decoder on ogbl-citation2 (Fig.~\ref{fig:per-edge-main}\subref{fig:per-edge-main-citation2-sign-gcn}). These observations show that the separation of ego- and neighbor-embeddings in the SAGE and SIGN encoders help GNNs to better adapt to local variations in feature similarity and node degrees in the real-world graphs.

\section{Conclusion}

We first characterized in this work the notion of non-homophilic link prediction based on the distributions of feature similarities between connected and unconnected nodes. We then proposed a theoretical framework that highlights the different optimizations needed for homophilic and heterophilic link prediction tasks. We further analyzed how different link prediction encoders and decoders adapt to varying levels of feature homophily, and identified designs---specifically, adopting learnable decoders such as MLP and DistMult with GNN encoders that separates ego- and neighbor-embedding in message passing---for improved overall performance and robustness to local feature similarity variations for link prediction tasks beyond homophily. 
Our experiments on synthetic and real-world datasets revealed the performance trend across feature similarity spectrum for various GNN4LP methods and demonstrated the effectiveness of our recommended designs. 

{
\small
\bibliographystyle{BIB/ACM-Reference-Format}
\bibliography{BIB/abbreviations,BIB/ACM-abbreviations,BIB/main,BIB/all}


 \providecommand{\noopsort}[1]{}
\begin{thebibliography}{55}


\ifx \showCODEN    \undefined \def \showCODEN     #1{\unskip}     \fi
\ifx \showDOI      \undefined \def \showDOI       #1{#1}\fi
\ifx \showISBNx    \undefined \def \showISBNx     #1{\unskip}     \fi
\ifx \showISBNxiii \undefined \def \showISBNxiii  #1{\unskip}     \fi
\ifx \showISSN     \undefined \def \showISSN      #1{\unskip}     \fi
\ifx \showLCCN     \undefined \def \showLCCN      #1{\unskip}     \fi
\ifx \shownote     \undefined \def \shownote      #1{#1}          \fi
\ifx \showarticletitle \undefined \def \showarticletitle #1{#1}   \fi
\ifx \showURL      \undefined \def \showURL       {\relax}        \fi
\providecommand\bibfield[2]{#2}
\providecommand\bibinfo[2]{#2}
\providecommand\natexlab[1]{#1}
\providecommand\showeprint[2][]{arXiv:#2}

\bibitem[\protect\citeauthoryear{Abu-El-Haija, Perozzi, Kapoor, Alipourfard, Lerman, Harutyunyan, Ver~Steeg, and Galstyan}{Abu-El-Haija et~al\mbox{.}}{2019}]%
        {abu2019mixhop}
\bibfield{author}{\bibinfo{person}{Sami Abu-El-Haija}, \bibinfo{person}{Bryan Perozzi}, \bibinfo{person}{Amol Kapoor}, \bibinfo{person}{Nazanin Alipourfard}, \bibinfo{person}{Kristina Lerman}, \bibinfo{person}{Hrayr Harutyunyan}, \bibinfo{person}{Greg Ver~Steeg}, {and} \bibinfo{person}{Aram Galstyan}.} \bibinfo{year}{2019}\natexlab{}.
\newblock \showarticletitle{Mixhop: Higher-order graph convolutional architectures via sparsified neighborhood mixing}. In \bibinfo{booktitle}{\emph{international conference on machine learning}}. PMLR, \bibinfo{pages}{21--29}.
\newblock


\bibitem[\protect\citeauthoryear{Adamic and Adar}{Adamic and Adar}{2003}]%
        {adamic2003friends}
\bibfield{author}{\bibinfo{person}{Lada~A Adamic} {and} \bibinfo{person}{Eytan Adar}.} \bibinfo{year}{2003}\natexlab{}.
\newblock \showarticletitle{Friends and neighbors on the web}.
\newblock \bibinfo{journal}{\emph{Social networks}} \bibinfo{volume}{25}, \bibinfo{number}{3} (\bibinfo{year}{2003}), \bibinfo{pages}{211--230}.
\newblock


\bibitem[\protect\citeauthoryear{Barab{\'a}si and Albert}{Barab{\'a}si and Albert}{1999}]%
        {barabasi1999emergence}
\bibfield{author}{\bibinfo{person}{Albert-L{\'a}szl{\'o} Barab{\'a}si} {and} \bibinfo{person}{R{\'e}ka Albert}.} \bibinfo{year}{1999}\natexlab{}.
\newblock \showarticletitle{Emergence of scaling in random networks}.
\newblock \bibinfo{journal}{\emph{science}} \bibinfo{volume}{286}, \bibinfo{number}{5439} (\bibinfo{year}{1999}), \bibinfo{pages}{509--512}.
\newblock


\bibitem[\protect\citeauthoryear{Bojchevski and G{\"u}nnemann}{Bojchevski and G{\"u}nnemann}{2017}]%
        {bojchevski2017deep}
\bibfield{author}{\bibinfo{person}{Aleksandar Bojchevski} {and} \bibinfo{person}{Stephan G{\"u}nnemann}.} \bibinfo{year}{2017}\natexlab{}.
\newblock \showarticletitle{Deep gaussian embedding of graphs: Unsupervised inductive learning via ranking}.
\newblock \bibinfo{journal}{\emph{arXiv preprint arXiv:1707.03815}} (\bibinfo{year}{2017}).
\newblock


\bibitem[\protect\citeauthoryear{Brin and Page}{Brin and Page}{1998}]%
        {brin1998anatomy}
\bibfield{author}{\bibinfo{person}{Sergey Brin} {and} \bibinfo{person}{Lawrence Page}.} \bibinfo{year}{1998}\natexlab{}.
\newblock \showarticletitle{The anatomy of a large-scale hypertextual web search engine}.
\newblock \bibinfo{journal}{\emph{Computer networks and ISDN systems}} \bibinfo{volume}{30}, \bibinfo{number}{1-7} (\bibinfo{year}{1998}), \bibinfo{pages}{107--117}.
\newblock


\bibitem[\protect\citeauthoryear{Chamberlain, Shirobokov, Rossi, Frasca, Markovich, Hammerla, Bronstein, and Hansmire}{Chamberlain et~al\mbox{.}}{2022}]%
        {chamberlain2022graph-ct}
\bibfield{author}{\bibinfo{person}{Benjamin~Paul Chamberlain}, \bibinfo{person}{Sergey Shirobokov}, \bibinfo{person}{Emanuele Rossi}, \bibinfo{person}{Fabrizio Frasca}, \bibinfo{person}{Thomas Markovich}, \bibinfo{person}{Nils~Yannick Hammerla}, \bibinfo{person}{Michael~M Bronstein}, {and} \bibinfo{person}{Max Hansmire}.} \bibinfo{year}{2022}\natexlab{}.
\newblock \showarticletitle{Graph neural networks for link prediction with subgraph sketching}. In \bibinfo{booktitle}{\emph{The eleventh international conference on learning representations}}.
\newblock


\bibitem[\protect\citeauthoryear{Daud, Ab~Hamid, Saadoon, Sahran, and Anuar}{Daud et~al\mbox{.}}{2020}]%
        {daud2020applications}
\bibfield{author}{\bibinfo{person}{Nur~Nasuha Daud}, \bibinfo{person}{Siti~Hafizah Ab~Hamid}, \bibinfo{person}{Muntadher Saadoon}, \bibinfo{person}{Firdaus Sahran}, {and} \bibinfo{person}{Nor~Badrul Anuar}.} \bibinfo{year}{2020}\natexlab{}.
\newblock \showarticletitle{Applications of link prediction in social networks: A review}.
\newblock \bibinfo{journal}{\emph{Journal of Network and Computer Applications}}  \bibinfo{volume}{166} (\bibinfo{year}{2020}), \bibinfo{pages}{102716}.
\newblock


\bibitem[\protect\citeauthoryear{Di~Francesco, Caso, Bucarelli, and Silvestri}{Di~Francesco et~al\mbox{.}}{2024}]%
        {di2024link}
\bibfield{author}{\bibinfo{person}{Andrea~Giuseppe Di~Francesco}, \bibinfo{person}{Francesco Caso}, \bibinfo{person}{Maria~Sofia Bucarelli}, {and} \bibinfo{person}{Fabrizio Silvestri}.} \bibinfo{year}{2024}\natexlab{}.
\newblock \showarticletitle{Link Prediction under Heterophily: A Physics-Inspired Graph Neural Network Approach}.
\newblock \bibinfo{journal}{\emph{arXiv preprint arXiv:2402.14802}} (\bibinfo{year}{2024}).
\newblock


\bibitem[\protect\citeauthoryear{Di~Giovanni, Rowbottom, Chamberlain, Markovich, and Bronstein}{Di~Giovanni et~al\mbox{.}}{2022}]%
        {di2022understanding}
\bibfield{author}{\bibinfo{person}{Francesco Di~Giovanni}, \bibinfo{person}{James Rowbottom}, \bibinfo{person}{Benjamin~P Chamberlain}, \bibinfo{person}{Thomas Markovich}, {and} \bibinfo{person}{Michael~M Bronstein}.} \bibinfo{year}{2022}\natexlab{}.
\newblock \showarticletitle{Understanding convolution on graphs via energies}.
\newblock \bibinfo{journal}{\emph{arXiv preprint arXiv:2206.10991}} (\bibinfo{year}{2022}).
\newblock


\bibitem[\protect\citeauthoryear{Dou, Shu, Xia, Yu, and Sun}{Dou et~al\mbox{.}}{2021}]%
        {dou2021user}
\bibfield{author}{\bibinfo{person}{Yingtong Dou}, \bibinfo{person}{Kai Shu}, \bibinfo{person}{Congying Xia}, \bibinfo{person}{Philip~S Yu}, {and} \bibinfo{person}{Lichao Sun}.} \bibinfo{year}{2021}\natexlab{}.
\newblock \showarticletitle{User preference-aware fake news detection}. In \bibinfo{booktitle}{\emph{Proceedings of the 44th international ACM SIGIR conference on research and development in information retrieval}}. \bibinfo{pages}{2051--2055}.
\newblock


\bibitem[\protect\citeauthoryear{Douze, Guzhva, Deng, Johnson, Szilvasy, Mazaré, Lomeli, Hosseini, and Jégou}{Douze et~al\mbox{.}}{2024}]%
        {douze2024faiss}
\bibfield{author}{\bibinfo{person}{Matthijs Douze}, \bibinfo{person}{Alexandr Guzhva}, \bibinfo{person}{Chengqi Deng}, \bibinfo{person}{Jeff Johnson}, \bibinfo{person}{Gergely Szilvasy}, \bibinfo{person}{Pierre-Emmanuel Mazaré}, \bibinfo{person}{Maria Lomeli}, \bibinfo{person}{Lucas Hosseini}, {and} \bibinfo{person}{Hervé Jégou}.} \bibinfo{year}{2024}\natexlab{}.
\newblock \showarticletitle{The Faiss library}.
\newblock  (\bibinfo{year}{2024}).
\newblock
\showeprint[arxiv]{cs.LG/2401.08281}


\bibitem[\protect\citeauthoryear{Dwivedi, Joshi, Luu, Laurent, Bengio, and Bresson}{Dwivedi et~al\mbox{.}}{2023}]%
        {dwivedi2023benchmarking}
\bibfield{author}{\bibinfo{person}{Vijay~Prakash Dwivedi}, \bibinfo{person}{Chaitanya~K Joshi}, \bibinfo{person}{Anh~Tuan Luu}, \bibinfo{person}{Thomas Laurent}, \bibinfo{person}{Yoshua Bengio}, {and} \bibinfo{person}{Xavier Bresson}.} \bibinfo{year}{2023}\natexlab{}.
\newblock \showarticletitle{Benchmarking graph neural networks}.
\newblock \bibinfo{journal}{\emph{Journal of Machine Learning Research}} \bibinfo{volume}{24}, \bibinfo{number}{43} (\bibinfo{year}{2023}), \bibinfo{pages}{1--48}.
\newblock


\bibitem[\protect\citeauthoryear{Halcrow, Mosoi, Ruth, and Perozzi}{Halcrow et~al\mbox{.}}{2020}]%
        {halcrow2020grale}
\bibfield{author}{\bibinfo{person}{Jonathan Halcrow}, \bibinfo{person}{Alexandru Mosoi}, \bibinfo{person}{Sam Ruth}, {and} \bibinfo{person}{Bryan Perozzi}.} \bibinfo{year}{2020}\natexlab{}.
\newblock \showarticletitle{Grale: Designing networks for graph learning}. In \bibinfo{booktitle}{\emph{Proceedings of the 26th ACM SIGKDD international conference on knowledge discovery \& data mining}}. \bibinfo{pages}{2523--2532}.
\newblock


\bibitem[\protect\citeauthoryear{Hamilton, Ying, and Leskovec}{Hamilton et~al\mbox{.}}{2017a}]%
        {hamilton2017inductive}
\bibfield{author}{\bibinfo{person}{Will Hamilton}, \bibinfo{person}{Zhitao Ying}, {and} \bibinfo{person}{Jure Leskovec}.} \bibinfo{year}{2017}\natexlab{a}.
\newblock \showarticletitle{Inductive representation learning on large graphs}. In \bibinfo{booktitle}{\emph{Advances in neural information processing systems (NeurIPS)}}. \bibinfo{pages}{1024--1034}.
\newblock


\bibitem[\protect\citeauthoryear{Hamilton, Ying, and Leskovec}{Hamilton et~al\mbox{.}}{2017b}]%
        {hamilton2017representation}
\bibfield{author}{\bibinfo{person}{William~L Hamilton}, \bibinfo{person}{Rex Ying}, {and} \bibinfo{person}{Jure Leskovec}.} \bibinfo{year}{2017}\natexlab{b}.
\newblock \showarticletitle{Representation learning on graphs: Methods and applications}.
\newblock \bibinfo{journal}{\emph{arXiv preprint arXiv:1709.05584}} (\bibinfo{year}{2017}).
\newblock


\bibitem[\protect\citeauthoryear{He, Wei, Xu, et~al\mbox{.}}{He et~al\mbox{.}}{2021}]%
        {he2021bernnet}
\bibfield{author}{\bibinfo{person}{Mingguo He}, \bibinfo{person}{Zhewei Wei}, \bibinfo{person}{Hongteng Xu}, {et~al\mbox{.}}} \bibinfo{year}{2021}\natexlab{}.
\newblock \showarticletitle{Bernnet: Learning arbitrary graph spectral filters via bernstein approximation}.
\newblock \bibinfo{journal}{\emph{Advances in Neural Information Processing Systems}}  \bibinfo{volume}{34} (\bibinfo{year}{2021}), \bibinfo{pages}{14239--14251}.
\newblock


\bibitem[\protect\citeauthoryear{Hu, Fey, Zitnik, Dong, Ren, Liu, Catasta, and Leskovec}{Hu et~al\mbox{.}}{2020}]%
        {hu2020ogb}
\bibfield{author}{\bibinfo{person}{Weihua Hu}, \bibinfo{person}{Matthias Fey}, \bibinfo{person}{Marinka Zitnik}, \bibinfo{person}{Yuxiao Dong}, \bibinfo{person}{Hongyu Ren}, \bibinfo{person}{Bowen Liu}, \bibinfo{person}{Michele Catasta}, {and} \bibinfo{person}{Jure Leskovec}.} \bibinfo{year}{2020}\natexlab{}.
\newblock \showarticletitle{Open Graph Benchmark: Datasets for Machine Learning on Graphs}.
\newblock \bibinfo{journal}{\emph{arXiv preprint arXiv:2005.00687}} (\bibinfo{year}{2020}).
\newblock


\bibitem[\protect\citeauthoryear{Kipf and Welling}{Kipf and Welling}{2016}]%
        {kipf2016variational}
\bibfield{author}{\bibinfo{person}{Thomas~N Kipf} {and} \bibinfo{person}{Max Welling}.} \bibinfo{year}{2016}\natexlab{}.
\newblock \showarticletitle{Variational graph auto-encoders}.
\newblock \bibinfo{journal}{\emph{arXiv preprint arXiv:1611.07308}} (\bibinfo{year}{2016}).
\newblock


\bibitem[\protect\citeauthoryear{Kipf and Welling}{Kipf and Welling}{2017}]%
        {kipf2016semi}
\bibfield{author}{\bibinfo{person}{Thomas~N. Kipf} {and} \bibinfo{person}{Max Welling}.} \bibinfo{year}{2017}\natexlab{}.
\newblock \showarticletitle{Semi-Supervised Classification with Graph Convolutional Networks}. In \bibinfo{booktitle}{\emph{International Conference on Learning Representations (ICLR)}}.
\newblock


\bibitem[\protect\citeauthoryear{Lee, Yu, Zhang, Ioannidis, song, Adeshina, Zheng, and Faloutsos}{Lee et~al\mbox{.}}{2024}]%
        {lee2024netinfof}
\bibfield{author}{\bibinfo{person}{Meng-Chieh Lee}, \bibinfo{person}{Haiyang Yu}, \bibinfo{person}{Jian Zhang}, \bibinfo{person}{Vassilis~N. Ioannidis}, \bibinfo{person}{Xiang song}, \bibinfo{person}{Soji Adeshina}, \bibinfo{person}{Da Zheng}, {and} \bibinfo{person}{Christos Faloutsos}.} \bibinfo{year}{2024}\natexlab{}.
\newblock \showarticletitle{NetInfoF Framework: Measuring and Exploiting Network Usable Information}. In \bibinfo{booktitle}{\emph{The Twelfth International Conference on Learning Representations}}.
\newblock
\urldef\tempurl%
\url{https://openreview.net/forum?id=KY8ZNcljVU}
\showURL{%
\tempurl}


\bibitem[\protect\citeauthoryear{Li, Shomer, Mao, Zeng, Ma, Shah, Tang, and Yin}{Li et~al\mbox{.}}{2023}]%
        {li2023evaluating-lb}
\bibfield{author}{\bibinfo{person}{Juanhui Li}, \bibinfo{person}{Harry Shomer}, \bibinfo{person}{Haitao Mao}, \bibinfo{person}{Shenglai Zeng}, \bibinfo{person}{Yao Ma}, \bibinfo{person}{Neil Shah}, \bibinfo{person}{Jiliang Tang}, {and} \bibinfo{person}{Dawei Yin}.} \bibinfo{year}{2023}\natexlab{}.
\newblock \showarticletitle{{Evaluating graph neural networks for link prediction: Current pitfalls and new benchmarking}}.
\newblock \bibinfo{journal}{\emph{arXiv [cs.LG]}} (\bibinfo{date}{June} \bibinfo{year}{2023}).
\newblock


\bibitem[\protect\citeauthoryear{Liang, Ding, Li, Liang, Wang, Chen, et~al\mbox{.}}{Liang et~al\mbox{.}}{2022}]%
        {liang2022can}
\bibfield{author}{\bibinfo{person}{Shuming Liang}, \bibinfo{person}{Yu Ding}, \bibinfo{person}{Zhidong Li}, \bibinfo{person}{Bin Liang}, \bibinfo{person}{Yang Wang}, \bibinfo{person}{Fang Chen}, {et~al\mbox{.}}} \bibinfo{year}{2022}\natexlab{}.
\newblock \showarticletitle{Can GNNs Learn Heuristic Information for Link Prediction?}
\newblock  (\bibinfo{year}{2022}).
\newblock


\bibitem[\protect\citeauthoryear{Lim, Hohne, Li, Huang, Gupta, Bhalerao, and Lim}{Lim et~al\mbox{.}}{2021}]%
        {lim2021new}
\bibfield{author}{\bibinfo{person}{Derek Lim}, \bibinfo{person}{Felix Hohne}, \bibinfo{person}{Xiuyu Li}, \bibinfo{person}{Sijia~Linda Huang}, \bibinfo{person}{Vaishnavi Gupta}, \bibinfo{person}{Omkar Bhalerao}, {and} \bibinfo{person}{Ser~Nam Lim}.} \bibinfo{year}{2021}\natexlab{}.
\newblock \showarticletitle{Large Scale Learning on Non-Homophilous Graphs: New Benchmarks and Strong Simple Methods}. In \bibinfo{booktitle}{\emph{NeurIPS}}.
\newblock


\bibitem[\protect\citeauthoryear{Luan, Hua, Lu, Zhu, Zhao, Zhang, Chang, and Precup}{Luan et~al\mbox{.}}{2022}]%
        {luan2022revisiting}
\bibfield{author}{\bibinfo{person}{Sitao Luan}, \bibinfo{person}{Chenqing Hua}, \bibinfo{person}{Qincheng Lu}, \bibinfo{person}{Jiaqi Zhu}, \bibinfo{person}{Mingde Zhao}, \bibinfo{person}{Shuyuan Zhang}, \bibinfo{person}{Xiao-Wen Chang}, {and} \bibinfo{person}{Doina Precup}.} \bibinfo{year}{2022}\natexlab{}.
\newblock \showarticletitle{Revisiting heterophily for graph neural networks}.
\newblock \bibinfo{journal}{\emph{Advances in neural information processing systems}}  \bibinfo{volume}{35} (\bibinfo{year}{2022}), \bibinfo{pages}{1362--1375}.
\newblock


\bibitem[\protect\citeauthoryear{Luan, Hua, Xu, Lu, Zhu, Chang, Fu, Leskovec, and Precup}{Luan et~al\mbox{.}}{2023}]%
        {luan2023when}
\bibfield{author}{\bibinfo{person}{Sitao Luan}, \bibinfo{person}{Chenqing Hua}, \bibinfo{person}{Minkai Xu}, \bibinfo{person}{Qincheng Lu}, \bibinfo{person}{Jiaqi Zhu}, \bibinfo{person}{Xiao-Wen Chang}, \bibinfo{person}{Jie Fu}, \bibinfo{person}{Jure Leskovec}, {and} \bibinfo{person}{Doina Precup}.} \bibinfo{year}{2023}\natexlab{}.
\newblock \showarticletitle{When Do Graph Neural Networks Help with Node Classification? Investigating the Homophily Principle on Node Distinguishability}. In \bibinfo{booktitle}{\emph{Thirty-seventh Conference on Neural Information Processing Systems}}.
\newblock
\urldef\tempurl%
\url{https://openreview.net/forum?id=kJmYu3Ti2z}
\showURL{%
\tempurl}


\bibitem[\protect\citeauthoryear{Ma, Liu, Shah, and Tang}{Ma et~al\mbox{.}}{2022}]%
        {ma2022is}
\bibfield{author}{\bibinfo{person}{Yao Ma}, \bibinfo{person}{Xiaorui Liu}, \bibinfo{person}{Neil Shah}, {and} \bibinfo{person}{Jiliang Tang}.} \bibinfo{year}{2022}\natexlab{}.
\newblock \showarticletitle{Is Homophily a Necessity for Graph Neural Networks?}. In \bibinfo{booktitle}{\emph{International Conference on Learning Representations}}.
\newblock
\urldef\tempurl%
\url{https://openreview.net/forum?id=ucASPPD9GKN}
\showURL{%
\tempurl}


\bibitem[\protect\citeauthoryear{Mao, Li, Shomer, Li, Fan, Ma, Zhao, Shah, and Tang}{Mao et~al\mbox{.}}{2023}]%
        {mao2023revisiting}
\bibfield{author}{\bibinfo{person}{Haitao Mao}, \bibinfo{person}{Juanhui Li}, \bibinfo{person}{Harry Shomer}, \bibinfo{person}{Bingheng Li}, \bibinfo{person}{Wenqi Fan}, \bibinfo{person}{Yao Ma}, \bibinfo{person}{Tong Zhao}, \bibinfo{person}{Neil Shah}, {and} \bibinfo{person}{Jiliang Tang}.} \bibinfo{year}{2023}\natexlab{}.
\newblock \showarticletitle{Revisiting link prediction: A data perspective}.
\newblock \bibinfo{journal}{\emph{arXiv preprint arXiv:2310.00793}} (\bibinfo{year}{2023}).
\newblock


\bibitem[\protect\citeauthoryear{Morris, Kriege, Bause, Kersting, Mutzel, and Neumann}{Morris et~al\mbox{.}}{2020}]%
        {morris2020tudataset}
\bibfield{author}{\bibinfo{person}{Christopher Morris}, \bibinfo{person}{Nils~M Kriege}, \bibinfo{person}{Franka Bause}, \bibinfo{person}{Kristian Kersting}, \bibinfo{person}{Petra Mutzel}, {and} \bibinfo{person}{Marion Neumann}.} \bibinfo{year}{2020}\natexlab{}.
\newblock \showarticletitle{Tudataset: A collection of benchmark datasets for learning with graphs}.
\newblock \bibinfo{journal}{\emph{arXiv preprint arXiv:2007.08663}} (\bibinfo{year}{2020}).
\newblock


\bibitem[\protect\citeauthoryear{Pei, Wei, Chang, Lei, and Yang}{Pei et~al\mbox{.}}{2020}]%
        {pei2020geom}
\bibfield{author}{\bibinfo{person}{Hongbin Pei}, \bibinfo{person}{Bingzhe Wei}, \bibinfo{person}{Kevin Chen-Chuan Chang}, \bibinfo{person}{Yu Lei}, {and} \bibinfo{person}{Bo Yang}.} \bibinfo{year}{2020}\natexlab{}.
\newblock \showarticletitle{Geom-gcn: Geometric graph convolutional networks}.
\newblock \bibinfo{journal}{\emph{arXiv preprint arXiv:2002.05287}} (\bibinfo{year}{2020}).
\newblock


\bibitem[\protect\citeauthoryear{Platonov, Kuznedelev, Babenko, and Prokhorenkova}{Platonov et~al\mbox{.}}{2024}]%
        {platonov2024characterizing}
\bibfield{author}{\bibinfo{person}{Oleg Platonov}, \bibinfo{person}{Denis Kuznedelev}, \bibinfo{person}{Artem Babenko}, {and} \bibinfo{person}{Liudmila Prokhorenkova}.} \bibinfo{year}{2024}\natexlab{}.
\newblock \showarticletitle{Characterizing graph datasets for node classification: Homophily-heterophily dichotomy and beyond}.
\newblock \bibinfo{journal}{\emph{Advances in Neural Information Processing Systems}}  \bibinfo{volume}{36} (\bibinfo{year}{2024}).
\newblock


\bibitem[\protect\citeauthoryear{Platonov, Kuznedelev, Diskin, Babenko, and Prokhorenkova}{Platonov et~al\mbox{.}}{2022}]%
        {platonov2022critical}
\bibfield{author}{\bibinfo{person}{Oleg Platonov}, \bibinfo{person}{Denis Kuznedelev}, \bibinfo{person}{Michael Diskin}, \bibinfo{person}{Artem Babenko}, {and} \bibinfo{person}{Liudmila Prokhorenkova}.} \bibinfo{year}{2022}\natexlab{}.
\newblock \showarticletitle{A critical look at the evaluation of GNNs under heterophily: Are we really making progress?}. In \bibinfo{booktitle}{\emph{The Eleventh International Conference on Learning Representations}}.
\newblock


\bibitem[\protect\citeauthoryear{Reddy, M{\`a}rquez, Valero, Rao, Zaragoza, Bandyopadhyay, Biswas, Xing, and Subbian}{Reddy et~al\mbox{.}}{2022}]%
        {reddy2022shopping}
\bibfield{author}{\bibinfo{person}{Chandan~K Reddy}, \bibinfo{person}{Llu{\'\i}s M{\`a}rquez}, \bibinfo{person}{Fran Valero}, \bibinfo{person}{Nikhil Rao}, \bibinfo{person}{Hugo Zaragoza}, \bibinfo{person}{Sambaran Bandyopadhyay}, \bibinfo{person}{Arnab Biswas}, \bibinfo{person}{Anlu Xing}, {and} \bibinfo{person}{Karthik Subbian}.} \bibinfo{year}{2022}\natexlab{}.
\newblock \showarticletitle{Shopping queries dataset: A large-scale ESCI benchmark for improving product search}.
\newblock \bibinfo{journal}{\emph{arXiv preprint arXiv:2206.06588}} (\bibinfo{year}{2022}).
\newblock


\bibitem[\protect\citeauthoryear{Rossi, Barbosa, Firmani, Matinata, and Merialdo}{Rossi et~al\mbox{.}}{2021}]%
        {rossi2021knowledge}
\bibfield{author}{\bibinfo{person}{Andrea Rossi}, \bibinfo{person}{Denilson Barbosa}, \bibinfo{person}{Donatella Firmani}, \bibinfo{person}{Antonio Matinata}, {and} \bibinfo{person}{Paolo Merialdo}.} \bibinfo{year}{2021}\natexlab{}.
\newblock \showarticletitle{Knowledge graph embedding for link prediction: A comparative analysis}.
\newblock \bibinfo{journal}{\emph{ACM Transactions on Knowledge Discovery from Data (TKDD)}} \bibinfo{volume}{15}, \bibinfo{number}{2} (\bibinfo{year}{2021}), \bibinfo{pages}{1--49}.
\newblock


\bibitem[\protect\citeauthoryear{Rozemberczki, Allen, and Sarkar}{Rozemberczki et~al\mbox{.}}{2021}]%
        {rozemberczki2021multi}
\bibfield{author}{\bibinfo{person}{Benedek Rozemberczki}, \bibinfo{person}{Carl Allen}, {and} \bibinfo{person}{Rik Sarkar}.} \bibinfo{year}{2021}\natexlab{}.
\newblock \showarticletitle{Multi-scale attributed node embedding}.
\newblock \bibinfo{journal}{\emph{Journal of Complex Networks}} \bibinfo{volume}{9}, \bibinfo{number}{2} (\bibinfo{year}{2021}), \bibinfo{pages}{cnab014}.
\newblock


\bibitem[\protect\citeauthoryear{Shchur, Mumme, Bojchevski, and G{\"u}nnemann}{Shchur et~al\mbox{.}}{2018}]%
        {shchur2018pitfalls}
\bibfield{author}{\bibinfo{person}{Oleksandr Shchur}, \bibinfo{person}{Maximilian Mumme}, \bibinfo{person}{Aleksandar Bojchevski}, {and} \bibinfo{person}{Stephan G{\"u}nnemann}.} \bibinfo{year}{2018}\natexlab{}.
\newblock \showarticletitle{Pitfalls of Graph Neural Network Evaluation}.
\newblock \bibinfo{journal}{\emph{Relational Representation Learning Workshop, NeurIPS 2018}} (\bibinfo{year}{2018}).
\newblock


\bibitem[\protect\citeauthoryear{Vahidi~Farashah, Etebarian, Azmi, and Ebrahimzadeh~Dastjerdi}{Vahidi~Farashah et~al\mbox{.}}{2021}]%
        {vahidi2021hybrid}
\bibfield{author}{\bibinfo{person}{Mohammadsadegh Vahidi~Farashah}, \bibinfo{person}{Akbar Etebarian}, \bibinfo{person}{Reza Azmi}, {and} \bibinfo{person}{Reza Ebrahimzadeh~Dastjerdi}.} \bibinfo{year}{2021}\natexlab{}.
\newblock \showarticletitle{A hybrid recommender system based-on link prediction for movie baskets analysis}.
\newblock \bibinfo{journal}{\emph{Journal of Big Data}}  \bibinfo{volume}{8} (\bibinfo{year}{2021}), \bibinfo{pages}{1--24}.
\newblock


\bibitem[\protect\citeauthoryear{Veli{\v{c}}kovi{\'{c}}, Cucurull, Casanova, Romero, Li{\`{o}}, and Bengio}{Veli{\v{c}}kovi{\'{c}} et~al\mbox{.}}{2018}]%
        {velickovic2018graph}
\bibfield{author}{\bibinfo{person}{Petar Veli{\v{c}}kovi{\'{c}}}, \bibinfo{person}{Guillem Cucurull}, \bibinfo{person}{Arantxa Casanova}, \bibinfo{person}{Adriana Romero}, \bibinfo{person}{Pietro Li{\`{o}}}, {and} \bibinfo{person}{Yoshua Bengio}.} \bibinfo{year}{2018}\natexlab{}.
\newblock \showarticletitle{{Graph Attention Networks}}.
\newblock \bibinfo{journal}{\emph{International Conference on Learning Representations (ICLR)}} (\bibinfo{year}{2018}).
\newblock
\urldef\tempurl%
\url{https://openreview.net/forum?id=rJXMpikCZ}
\showURL{%
\tempurl}


\bibitem[\protect\citeauthoryear{Wang, Yang, and Zhang}{Wang et~al\mbox{.}}{2023}]%
        {wang2023neural}
\bibfield{author}{\bibinfo{person}{Xiyuan Wang}, \bibinfo{person}{Haotong Yang}, {and} \bibinfo{person}{Muhan Zhang}.} \bibinfo{year}{2023}\natexlab{}.
\newblock \showarticletitle{Neural common neighbor with completion for link prediction}.
\newblock \bibinfo{journal}{\emph{arXiv preprint arXiv:2302.00890}} (\bibinfo{year}{2023}).
\newblock


\bibitem[\protect\citeauthoryear{Wang, Hooi, Liu, Zhao, Guo, and Shah}{Wang et~al\mbox{.}}{2022}]%
        {wang2022flashlight}
\bibfield{author}{\bibinfo{person}{Yiwei Wang}, \bibinfo{person}{Bryan Hooi}, \bibinfo{person}{Yozen Liu}, \bibinfo{person}{Tong Zhao}, \bibinfo{person}{Zhichun Guo}, {and} \bibinfo{person}{Neil Shah}.} \bibinfo{year}{2022}\natexlab{}.
\newblock \showarticletitle{Flashlight: Scalable link prediction with effective decoders}. In \bibinfo{booktitle}{\emph{Learning on Graphs Conference}}. PMLR, \bibinfo{pages}{14--1}.
\newblock


\bibitem[\protect\citeauthoryear{Wu, Ramsundar, Feinberg, Gomes, Geniesse, Pappu, Leswing, and Pande}{Wu et~al\mbox{.}}{2018}]%
        {wu2018moleculenet}
\bibfield{author}{\bibinfo{person}{Zhenqin Wu}, \bibinfo{person}{Bharath Ramsundar}, \bibinfo{person}{Evan~N Feinberg}, \bibinfo{person}{Joseph Gomes}, \bibinfo{person}{Caleb Geniesse}, \bibinfo{person}{Aneesh~S Pappu}, \bibinfo{person}{Karl Leswing}, {and} \bibinfo{person}{Vijay Pande}.} \bibinfo{year}{2018}\natexlab{}.
\newblock \showarticletitle{MoleculeNet: a benchmark for molecular machine learning}.
\newblock \bibinfo{journal}{\emph{Chemical science}} \bibinfo{volume}{9}, \bibinfo{number}{2} (\bibinfo{year}{2018}), \bibinfo{pages}{513--530}.
\newblock


\bibitem[\protect\citeauthoryear{Yang, Yih, He, Gao, and Deng}{Yang et~al\mbox{.}}{2014}]%
        {yang2014embedding}
\bibfield{author}{\bibinfo{person}{Bishan Yang}, \bibinfo{person}{Wen-tau Yih}, \bibinfo{person}{Xiaodong He}, \bibinfo{person}{Jianfeng Gao}, {and} \bibinfo{person}{Li Deng}.} \bibinfo{year}{2014}\natexlab{}.
\newblock \showarticletitle{Embedding entities and relations for learning and inference in knowledge bases}.
\newblock \bibinfo{journal}{\emph{arXiv preprint arXiv:1412.6575}} (\bibinfo{year}{2014}).
\newblock


\bibitem[\protect\citeauthoryear{Yang, Shi, Xiao, Yang, Bhowmick, and Liu}{Yang et~al\mbox{.}}{2023}]%
        {yang2023pane}
\bibfield{author}{\bibinfo{person}{Renchi Yang}, \bibinfo{person}{Jieming Shi}, \bibinfo{person}{Xiaokui Xiao}, \bibinfo{person}{Yin Yang}, \bibinfo{person}{Sourav~S Bhowmick}, {and} \bibinfo{person}{Juncheng Liu}.} \bibinfo{year}{2023}\natexlab{}.
\newblock \showarticletitle{PANE: scalable and effective attributed network embedding}.
\newblock \bibinfo{journal}{\emph{The VLDB Journal}} \bibinfo{volume}{32}, \bibinfo{number}{6} (\bibinfo{year}{2023}), \bibinfo{pages}{1237--1262}.
\newblock


\bibitem[\protect\citeauthoryear{Yun, Kim, Lee, Kang, and Kim}{Yun et~al\mbox{.}}{2022}]%
        {yun2022neo-gnns-cr}
\bibfield{author}{\bibinfo{person}{Seongjun Yun}, \bibinfo{person}{Seoyoon Kim}, \bibinfo{person}{Junhyun Lee}, \bibinfo{person}{Jaewoo Kang}, {and} \bibinfo{person}{Hyunwoo~J Kim}.} \bibinfo{year}{2022}\natexlab{}.
\newblock \showarticletitle{{Neo-GNNs: Neighborhood Overlap-aware Graph Neural Networks for link prediction}}.
\newblock \bibinfo{journal}{\emph{arXiv [cs.LG]}} (\bibinfo{date}{June} \bibinfo{year}{2022}).
\newblock


\bibitem[\protect\citeauthoryear{Zachary}{Zachary}{1977}]%
        {zachary1977information}
\bibfield{author}{\bibinfo{person}{Wayne~W Zachary}.} \bibinfo{year}{1977}\natexlab{}.
\newblock \showarticletitle{An information flow model for conflict and fission in small groups}.
\newblock \bibinfo{journal}{\emph{Journal of anthropological research}} \bibinfo{volume}{33}, \bibinfo{number}{4} (\bibinfo{year}{1977}), \bibinfo{pages}{452--473}.
\newblock


\bibitem[\protect\citeauthoryear{Zhang}{Zhang}{2022}]%
        {zhang2022graph}
\bibfield{author}{\bibinfo{person}{Muhan Zhang}.} \bibinfo{year}{2022}\natexlab{}.
\newblock \showarticletitle{Graph neural networks: link prediction}.
\newblock \bibinfo{journal}{\emph{Graph Neural Networks: Foundations, Frontiers, and Applications}} (\bibinfo{year}{2022}), \bibinfo{pages}{195--223}.
\newblock


\bibitem[\protect\citeauthoryear{Zhang and Chen}{Zhang and Chen}{2018}]%
        {zhang2018link}
\bibfield{author}{\bibinfo{person}{Muhan Zhang} {and} \bibinfo{person}{Yixin Chen}.} \bibinfo{year}{2018}\natexlab{}.
\newblock \showarticletitle{Link prediction based on graph neural networks}.
\newblock \bibinfo{journal}{\emph{Advances in neural information processing systems}}  \bibinfo{volume}{31} (\bibinfo{year}{2018}).
\newblock


\bibitem[\protect\citeauthoryear{Zhang, Li, Xia, Wang, and Jin}{Zhang et~al\mbox{.}}{2020}]%
        {zhang2020revisiting}
\bibfield{author}{\bibinfo{person}{Muhan Zhang}, \bibinfo{person}{Pan Li}, \bibinfo{person}{Yinglong Xia}, \bibinfo{person}{Kai Wang}, {and} \bibinfo{person}{Long Jin}.} \bibinfo{year}{2020}\natexlab{}.
\newblock \showarticletitle{Revisiting graph neural networks for link prediction}.
\newblock  (\bibinfo{year}{2020}).
\newblock


\bibitem[\protect\citeauthoryear{Zhang, Li, Xia, Wang, and Jin}{Zhang et~al\mbox{.}}{2021}]%
        {zhang2021labeling}
\bibfield{author}{\bibinfo{person}{Muhan Zhang}, \bibinfo{person}{Pan Li}, \bibinfo{person}{Yinglong Xia}, \bibinfo{person}{Kai Wang}, {and} \bibinfo{person}{Long Jin}.} \bibinfo{year}{2021}\natexlab{}.
\newblock \showarticletitle{Labeling trick: A theory of using graph neural networks for multi-node representation learning}.
\newblock \bibinfo{journal}{\emph{Advances in Neural Information Processing Systems}}  \bibinfo{volume}{34} (\bibinfo{year}{2021}), \bibinfo{pages}{9061--9073}.
\newblock


\bibitem[\protect\citeauthoryear{Zheng, Wang, Liu, Li, Zhang, Jin, Yu, and Pan}{Zheng et~al\mbox{.}}{2022}]%
        {zheng2022graph}
\bibfield{author}{\bibinfo{person}{Xin Zheng}, \bibinfo{person}{Yi Wang}, \bibinfo{person}{Yixin Liu}, \bibinfo{person}{Ming Li}, \bibinfo{person}{Miao Zhang}, \bibinfo{person}{Di Jin}, \bibinfo{person}{Philip~S Yu}, {and} \bibinfo{person}{Shirui Pan}.} \bibinfo{year}{2022}\natexlab{}.
\newblock \showarticletitle{Graph neural networks for graphs with heterophily: A survey}.
\newblock \bibinfo{journal}{\emph{arXiv preprint arXiv:2202.07082}} (\bibinfo{year}{2022}).
\newblock


\bibitem[\protect\citeauthoryear{Zhou, Guo, Aggarwal, Zhang, and Wang}{Zhou et~al\mbox{.}}{2022}]%
        {zhou2022link}
\bibfield{author}{\bibinfo{person}{Shijie Zhou}, \bibinfo{person}{Zhimeng Guo}, \bibinfo{person}{Charu Aggarwal}, \bibinfo{person}{Xiang Zhang}, {and} \bibinfo{person}{Suhang Wang}.} \bibinfo{year}{2022}\natexlab{}.
\newblock \showarticletitle{Link prediction on heterophilic graphs via disentangled representation learning}.
\newblock \bibinfo{journal}{\emph{arXiv preprint arXiv:2208.01820}} (\bibinfo{year}{2022}).
\newblock


\bibitem[\protect\citeauthoryear{Zhu, Rossi, Rao, Mai, Lipka, Ahmed, and Koutra}{Zhu et~al\mbox{.}}{2020a}]%
        {zhu2020graph}
\bibfield{author}{\bibinfo{person}{Jiong Zhu}, \bibinfo{person}{Ryan~A Rossi}, \bibinfo{person}{Anup Rao}, \bibinfo{person}{Tung Mai}, \bibinfo{person}{Nedim Lipka}, \bibinfo{person}{Nesreen~K Ahmed}, {and} \bibinfo{person}{Danai Koutra}.} \bibinfo{year}{2020}\natexlab{a}.
\newblock \showarticletitle{Graph Neural Networks with Heterophily}.
\newblock \bibinfo{journal}{\emph{arXiv preprint arXiv:2009.13566}} (\bibinfo{year}{2020}).
\newblock


\bibitem[\protect\citeauthoryear{Zhu, Yan, Zhao, Heimann, Akoglu, and Koutra}{Zhu et~al\mbox{.}}{2020b}]%
        {zhu2020beyond}
\bibfield{author}{\bibinfo{person}{Jiong Zhu}, \bibinfo{person}{Yujun Yan}, \bibinfo{person}{Lingxiao Zhao}, \bibinfo{person}{Mark Heimann}, \bibinfo{person}{Leman Akoglu}, {and} \bibinfo{person}{Danai Koutra}.} \bibinfo{year}{2020}\natexlab{b}.
\newblock \showarticletitle{Beyond homophily in graph neural networks: Current limitations and effective designs}.
\newblock \bibinfo{journal}{\emph{Advances in neural information processing systems}}  \bibinfo{volume}{33} (\bibinfo{year}{2020}), \bibinfo{pages}{7793--7804}.
\newblock


\bibitem[\protect\citeauthoryear{Zhu, Zhou, Ioannidis, Qian, Ai, Song, and Koutra}{Zhu et~al\mbox{.}}{2024}]%
        {zhu2024pitfalls}
\bibfield{author}{\bibinfo{person}{Jing Zhu}, \bibinfo{person}{Yuhang Zhou}, \bibinfo{person}{Vassilis~N Ioannidis}, \bibinfo{person}{Shengyi Qian}, \bibinfo{person}{Wei Ai}, \bibinfo{person}{Xiang Song}, {and} \bibinfo{person}{Danai Koutra}.} \bibinfo{year}{2024}\natexlab{}.
\newblock \showarticletitle{Pitfalls in Link Prediction with Graph Neural Networks: Understanding the Impact of Target-link Inclusion \& Better Practices}. In \bibinfo{booktitle}{\emph{Proceedings of the 17th ACM International Conference on Web Search and Data Mining}}. \bibinfo{pages}{994--1002}.
\newblock


\bibitem[\protect\citeauthoryear{Zhu, Zhang, Xhonneux, and Tang}{Zhu et~al\mbox{.}}{2021}]%
        {zhu2021neural}
\bibfield{author}{\bibinfo{person}{Zhaocheng Zhu}, \bibinfo{person}{Zuobai Zhang}, \bibinfo{person}{Louis-Pascal Xhonneux}, {and} \bibinfo{person}{Jian Tang}.} \bibinfo{year}{2021}\natexlab{}.
\newblock \showarticletitle{Neural bellman-ford networks: A general graph neural network framework for link prediction}.
\newblock \bibinfo{journal}{\emph{Advances in Neural Information Processing Systems}}  \bibinfo{volume}{34} (\bibinfo{year}{2021}), \bibinfo{pages}{29476--29490}.
\newblock


\bibitem[\protect\citeauthoryear{Zitnik and Leskovec}{Zitnik and Leskovec}{2017}]%
        {zitnik2017predicting}
\bibfield{author}{\bibinfo{person}{Marinka Zitnik} {and} \bibinfo{person}{Jure Leskovec}.} \bibinfo{year}{2017}\natexlab{}.
\newblock \showarticletitle{Predicting multicellular function through multi-layer tissue networks}.
\newblock \bibinfo{journal}{\emph{Bioinformatics}} \bibinfo{volume}{33}, \bibinfo{number}{14} (\bibinfo{year}{2017}), \bibinfo{pages}{i190--i198}.
\newblock


\end{thebibliography}
}

\appendix
\newpage

\section{Additional Details on Experiments}
\label{app:sec:exp-details}

\paragraph{Computing Resources} For most experiments, we use a workstation with a 12-core AMD Ryzen 9 3900X CPU, 64GB RAM, and an NVIDIA Quadro P6000 GPU with 24 GB GPU Memory. BUDDY experiments on ogbl-citation2 require higher CPU RAM, in which case we use a server with 128GB RAM and an NVIDIA A100 GPU with 48 GB GPU Memory. 

\textbf{Synthetic Graph Generation.} 
The synthetic graphs are generated by random sampling 10,000 nodes with their features in ogbl-collab~\cite{hu2020ogb} and connect 2\% of all possible node pairs whose feature similarity falls within specified ranges; all graphs share the same set of nodes and features and only differ in their edges. 
More specifically, we calculate the pairwise feature similarity between all node pairs and create 50-quantiles of feature similarity scores. We select the 3 smallest quantiles, the 3 largest quantiles, and 4 quantiles in equal intervals in between, resulting in 10 quantiles. We then create 10 synthetic graphs by connecting node pairs whose feature similarity scores fall within the same quantile.
Thus, by gradually increasing the range of similarity for connected nodes, we create graphs which resemble different types of link prediction tasks and average feature similarity. 
We list the average feature similarity score of each synthetic graph in Table~\ref{fig:synthetic-results}: the negative extreme is the most negatively correlated graph featuring edges with feature similarity scores ranging in $[-1, -0.33]$, 
which resembles the heterophilic link prediction task; the positive extreme is the most positively correlated graph with edges similarity scores in $[0.44, 1.00]$, resembling the homophilic link prediction task. Other synthetic graphs correspond to the gated link prediction task with feature similarity scores ranging from the negative to positive spectrum. 

\textbf{Experiment Setups of the Synthetic Graphs.}  
For each generated synthetic graph, we randomly split the edges into training, validation, and test sets with a ratio of 8:1:1. We repeat each experiment 3 times with different random seeds and report the average performance with standard deviation in Table~\ref{tab:synthetic-results}. We use 2 convolutional layers for SAGE and GCN, and 256 hidden dimensions for all neural network models.

\paragraph{Additional Experiment Setups of the Real-world Datasets} We consider three real-world datasets: (1) ogbl-collab~\cite{hu2020ogb}, a collaboration network between researchers, with nodes representing authors, edges representing co-authorship, and node features as the average word embeddings of the author's papers; (2) ogbl-citation2~\cite{hu2020ogb}, a citation network where nodes represent papers and edges represent citations, with the average word embeddings of the paper's title and abstract as node features; (3) e-comm~\cite{zhu2024pitfalls}, a sparse graph extracted from \cite{reddy2022shopping} representing exact matches of queries and related products in Amazon Search, with BERT-embeddings of queries and product information as node features. For ogbl-collab and ogbl-citation2, we follow the recommended metrics (Hit@50 and MRR, respectively) and train-validation-test splits provided by OGB~\cite{hu2020ogb}. 
For e-comm, we use the splits shared by \cite{zhu2024pitfalls}, adopt the SpotTarget approach~\cite{zhu2024pitfalls} as graph mini-batch sampler for training and 
use MRR as the evaluation metric. We report each method's average performance on the full test split across 3 runs with different random seeds, and present the results in Table~\ref{tab:real-world-results}. 

\paragraph{Creation of Node Degree and Feature Similarity Buckets on Real-world Datsets} To create these buckets, we first calculate each edge's feature similarity score and the minimum degree of its two connected nodes. We then create on each graph three buckets per property based on the distribution of feature similarity scores and node degrees, with each bucket covering one-third quantile, except for node degrees of e-comm, where only two buckets are created due to its sparsity.


\begin{table}
    \small
    \centering
    \caption{Results on synthetic graphs. We report MRR averaged over 3 runs.}
    \label{tab:synthetic-results}
    \begin{tabular}{rccccccccccc}
		\toprule 
        \textbf{Graph Index} & \textbf{0} & \textbf{1} & \textbf{2} & \textbf{3} & \textbf{4}  \\
		\textbf{Feat. Sim $K$} & $-0.38{\scriptstyle \pm0.04}$ & $-0.31{\scriptstyle \pm0.01}$ & $-0.28{\scriptstyle \pm0.01}$ & $-0.14{\scriptstyle \pm0.00}$ & $-0.04{\scriptstyle \pm0.00}$ \\
        \midrule
        \textbf{CN} & $0.69$ & $0.65$ & $0.66$ & $1.58$ & $2.44$ \\
        \textbf{RA} & $0.69$ & $0.65$ & $0.66$ & $1.64$ & $2.58$\\
        \textbf{AA} & $0.69$ & $0.65$ & $0.66$ & $1.63$ & $2.57$\\
        \textbf{PPR} & $10.91$ & $3.30$ & $2.46$ & $3.17$ & $3.68$\\
        \midrule
        \textbf{GCN+DOT} & $15.88{\scriptstyle \pm0.37}$ & $8.16{\scriptstyle \pm0.11}$ & $5.96{\scriptstyle \pm0.07}$ & $3.55{\scriptstyle \pm0.02}$ & $3.79{\scriptstyle \pm0.00}$ \\
        \textbf{SAGE+DOT} & $25.53{\scriptstyle \pm0.09}$ & $9.35{\scriptstyle \pm0.02}$ & $6.32{\scriptstyle \pm0.02}$ & $3.80{\scriptstyle \pm0.07}$ & $3.51{\scriptstyle \pm0.04}$ \\
        \midrule
        \textbf{GCN+DistMult} & $46.19{\scriptstyle \pm0.98}$ & $17.20{\scriptstyle \pm0.16}$ & $12.57{\scriptstyle \pm0.05}$ & $4.86{\scriptstyle \pm0.03}$ & $3.58{\scriptstyle \pm0.11}$ \\
        \textbf{SAGE+DistMult} & $80.97{\scriptstyle \pm0.32}$ & $23.85{\scriptstyle \pm0.04}$ & $14.66{\scriptstyle \pm0.03}$ & $6.05{\scriptstyle \pm0.08}$ & $6.57{\scriptstyle \pm0.05}$ \\
        \midrule
        \textbf{NoGNN+MLP} & $57.84{\scriptstyle \pm0.37}$ & $23.83{\scriptstyle \pm0.08}$ & $18.30{\scriptstyle \pm0.09}$ & $9.05{\scriptstyle \pm0.02}$ & $7.89{\scriptstyle \pm0.04}$ \\
        \textbf{GCN+MLP} & $45.69{\scriptstyle \pm0.47}$ & $17.43{\scriptstyle \pm0.08}$ & $12.62{\scriptstyle \pm0.08}$ & $5.03{\scriptstyle \pm0.06}$ & $3.98{\scriptstyle \pm0.06}$ \\
        \textbf{SAGE+MLP} & $75.39{\scriptstyle \pm0.64}$ & $43.37{\scriptstyle \pm2.35}$ & $33.64{\scriptstyle \pm1.01}$ & $19.85{\scriptstyle \pm0.68}$ & $23.84{\scriptstyle \pm3.15}$  \\
        \midrule

        \textbf{BUDDY-NoFeat} & $27.51{\scriptstyle \pm0.05}$ & $10.12{\scriptstyle \pm0.95}$ & $6.72{\scriptstyle \pm0.85}$ & $2.37{\scriptstyle \pm0.37}$ & $2.15{\scriptstyle \pm0.11}$ \\
        \textbf{BUDDY-GCN} & $69.47{\scriptstyle \pm0.65}$ & $24.26{\scriptstyle \pm0.56}$ & $17.60{\scriptstyle \pm0.43}$ & $6.46{\scriptstyle \pm0.10}$ & $4.99{\scriptstyle \pm0.07}$ \\
        \textbf{BUDDY-SIGN} & $75.27{\scriptstyle \pm2.33}$ & $26.65{\scriptstyle \pm0.96}$ & $20.81{\scriptstyle \pm0.39}$ & $9.59{\scriptstyle \pm0.29}$ & $7.57{\scriptstyle \pm0.05}$ \\
        \bottomrule
        
        \toprule 
        \textbf{Graph Index} &  \textbf{5} & \textbf{6} & \textbf{7} & \textbf{8} & \textbf{9} \\
		\textbf{Feat Sim $K$} & $0.07{\scriptstyle \pm0.00}$ & $0.24{\scriptstyle \pm0.01}$ & $0.34{\scriptstyle \pm0.01}$ & $0.40{\scriptstyle \pm0.02}$ & $0.54{\scriptstyle \pm0.09}$ \\
        \midrule
        \textbf{CN} & 2.81& 6.26&11.97&18.62&70.49 \\
        \textbf{RA} & 2.96& 6.48&12.61&20.24&77.16\\
        \textbf{AA} & 2.96& 6.50&12.37&19.21&71.91\\
        \textbf{PPR} & 3.86& 7.44&13.21&19.62&62.06\\
        \midrule
        \textbf{GCN+DOT} & $4.16{\scriptstyle \pm0.04}$ & $8.11{\scriptstyle \pm0.02}$ & $14.27{\scriptstyle \pm0.05}$ & $20.23{\scriptstyle \pm0.10}$ & $61.57{\scriptstyle \pm0.22}$ \\
        \textbf{SAGE+DOT} & $2.29{\scriptstyle \pm0.10}$ & $6.95{\scriptstyle \pm0.08}$ & $14.79{\scriptstyle \pm0.02}$ & $23.95{\scriptstyle \pm0.02}$ & $79.90{\scriptstyle \pm0.13}$ \\
        \midrule
        \textbf{GCN+DistMult} & $4.08{\scriptstyle \pm0.01}$ & $7.97{\scriptstyle \pm0.01}$ & $13.99{\scriptstyle \pm0.03}$ & $19.85{\scriptstyle \pm0.08}$ & $62.92{\scriptstyle \pm0.82}$ \\
        \textbf{SAGE+DistMult} & $6.71{\scriptstyle \pm0.13}$ & $7.33{\scriptstyle \pm0.08}$ & $16.05{\scriptstyle \pm0.02}$ & $27.52{\scriptstyle \pm0.05}$ & $88.97{\scriptstyle \pm0.35}$ \\
        \midrule
        \textbf{NoGNN+MLP} & $8.86{\scriptstyle \pm0.13}$ & $14.78{\scriptstyle \pm0.21}$ & $23.59{\scriptstyle \pm0.32}$ & $31.88{\scriptstyle \pm0.55}$ & $70.55{\scriptstyle \pm0.83}$ \\
        \textbf{GCN+MLP} & $4.19{\scriptstyle \pm0.02}$ & $8.12{\scriptstyle \pm0.05}$ & $14.03{\scriptstyle \pm0.02}$ & $20.07{\scriptstyle \pm0.14}$ & $62.24{\scriptstyle \pm0.39}$ \\
        \textbf{SAGE+MLP} & $30.35{\scriptstyle \pm0.12}$ & $35.33{\scriptstyle \pm0.75}$ & $49.73{\scriptstyle \pm1.42}$ & $64.38{\scriptstyle \pm0.88}$ & $89.79{\scriptstyle \pm0.74}$ \\
        \midrule
        \textbf{BUDDY-NoFeat} & $2.30{\scriptstyle \pm0.08}$ & $5.10{\scriptstyle \pm0.97}$ & $10.54{\scriptstyle \pm2.05}$ & $18.13{\scriptstyle \pm0.04}$ & $67.96{\scriptstyle \pm0.01}$ \\
        \textbf{BUDDY-GCN} & $5.39{\scriptstyle \pm0.05}$ & $10.11{\scriptstyle \pm0.45}$ & $19.34{\scriptstyle \pm1.47}$ & $29.44{\scriptstyle \pm2.36}$ & $84.44{\scriptstyle \pm1.22}$ \\
        \textbf{BUDDY-SIGN} & $8.45{\scriptstyle \pm0.08}$ & $16.17{\scriptstyle \pm0.44}$ & $27.49{\scriptstyle \pm2.33}$ & $38.39{\scriptstyle \pm4.77}$ & $87.59{\scriptstyle \pm1.40}$ \\
        \bottomrule
    \end{tabular}
\end{table}

\begin{figure}
    \begin{subfigure}{0.32\textwidth}
        \includegraphics[width=\textwidth]{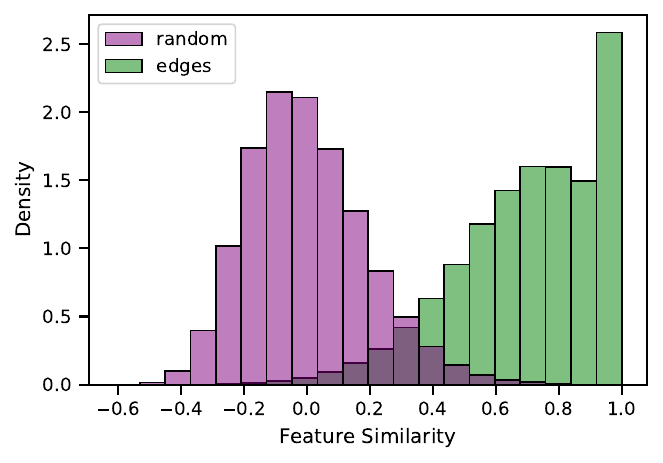}
        \caption{ogbl-collab}
    \end{subfigure}
    ~
    \begin{subfigure}{0.32\textwidth}
        \includegraphics[width=\textwidth]{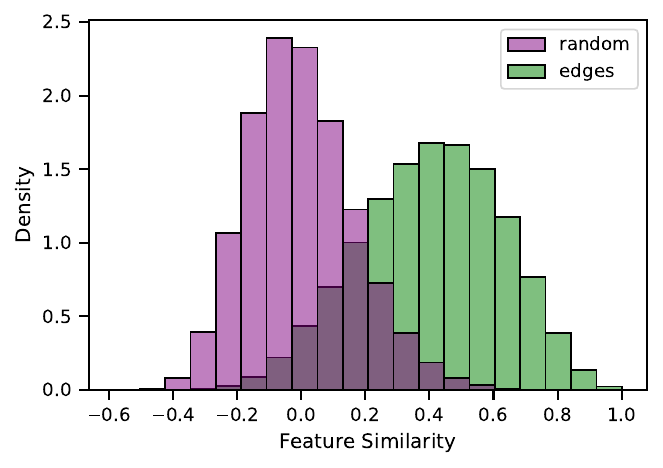}
        \caption{ogbl-citation2}
    \end{subfigure}
    ~
    \begin{subfigure}{0.32\textwidth}
        \includegraphics[width=\textwidth]{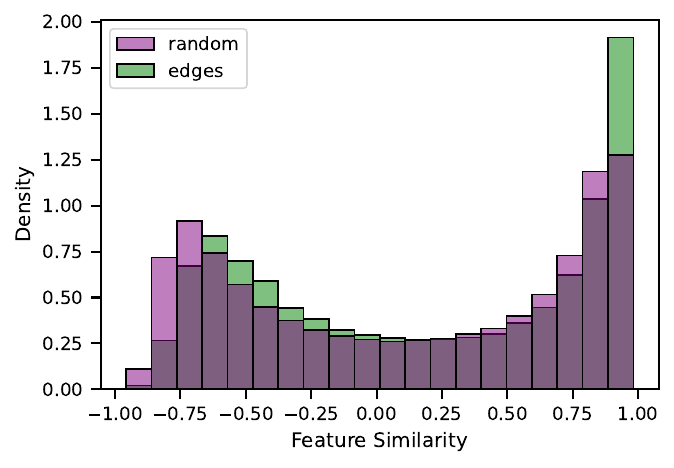}
        \caption{e-comm}
    \end{subfigure}
    \caption{Comparison of feature similarity distributions for edges and random node pairs on real-world datasets used in our experiments. For similarity scores of random node pairs, we randomly sample 1000 nodes and compute the pairwise cosine similarity between these node features. Similarity score distributions for random node pairs are good approximations of the distributions for non-edge node pairs due to the sparsity of the graphs.}
    \label{fig:real-graph-sim-hist}
\end{figure}

\begin{figure}
    \begin{subfigure}{0.32\textwidth}
        \includegraphics[width=\textwidth]{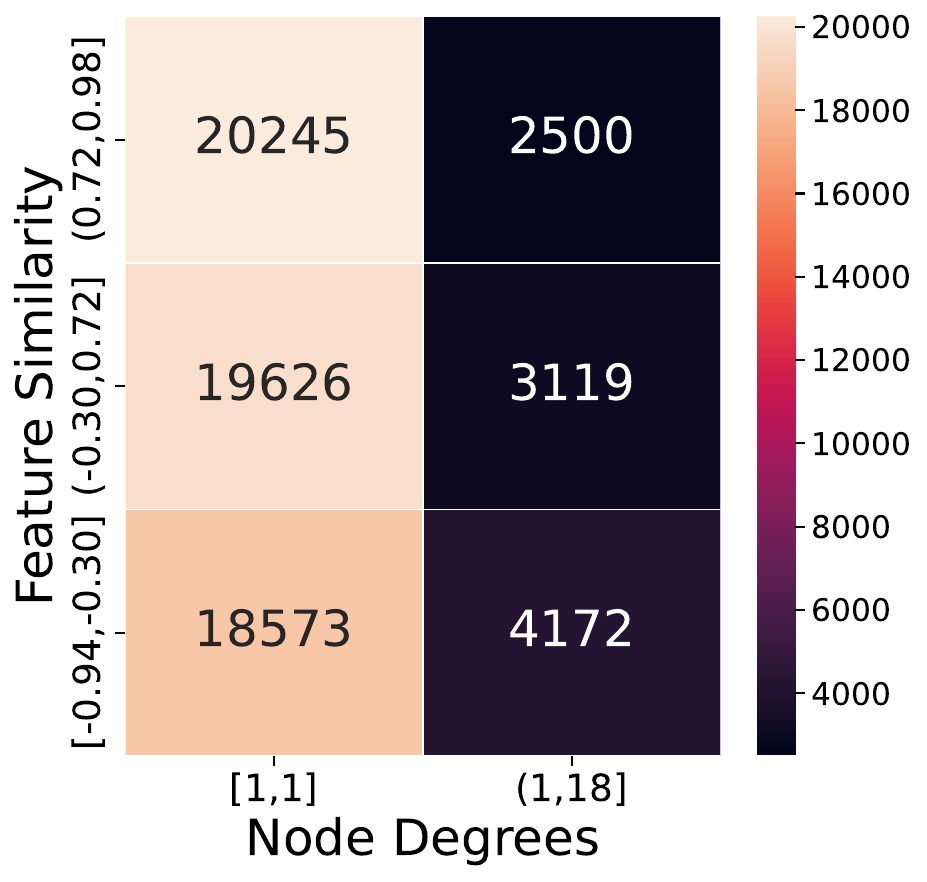}
        \caption{Number of test edges in each bucket on ogbl-collab.}
    \end{subfigure}
    ~
    \begin{subfigure}{0.32\textwidth}
        \includegraphics[width=\textwidth]{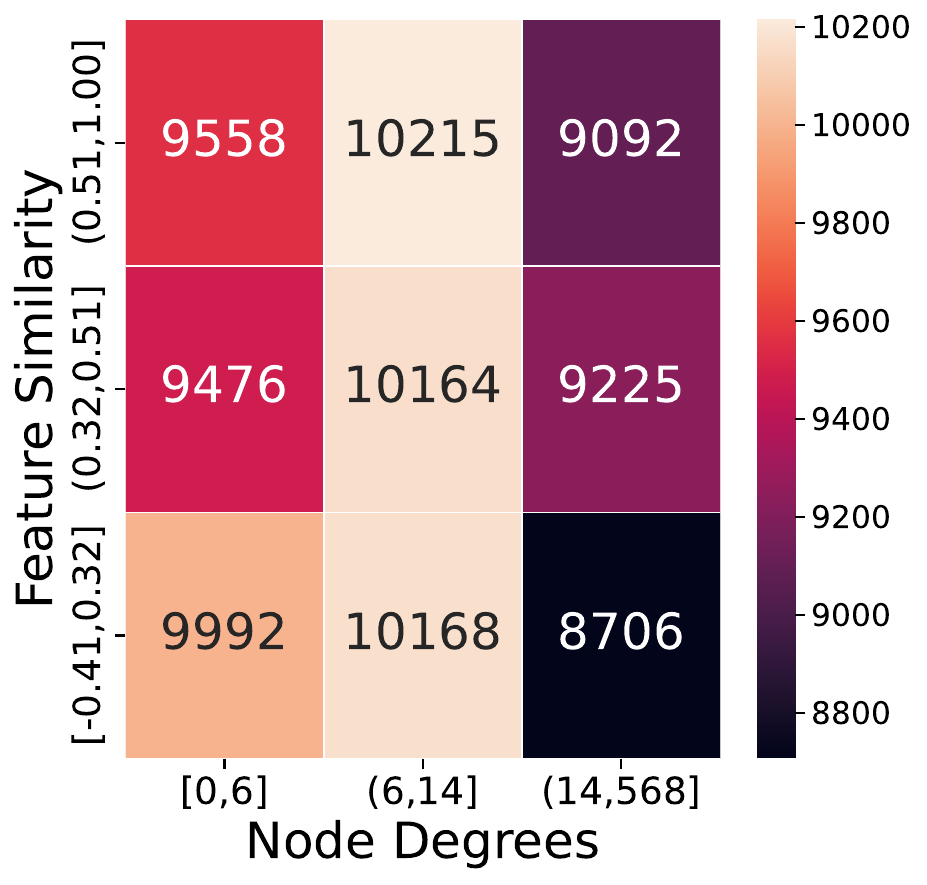}
        \caption{Number of test edges in each bucket on ogbl-citation2.}
    \end{subfigure}
    ~
    \begin{subfigure}{0.32\textwidth}
        \includegraphics[width=\textwidth]{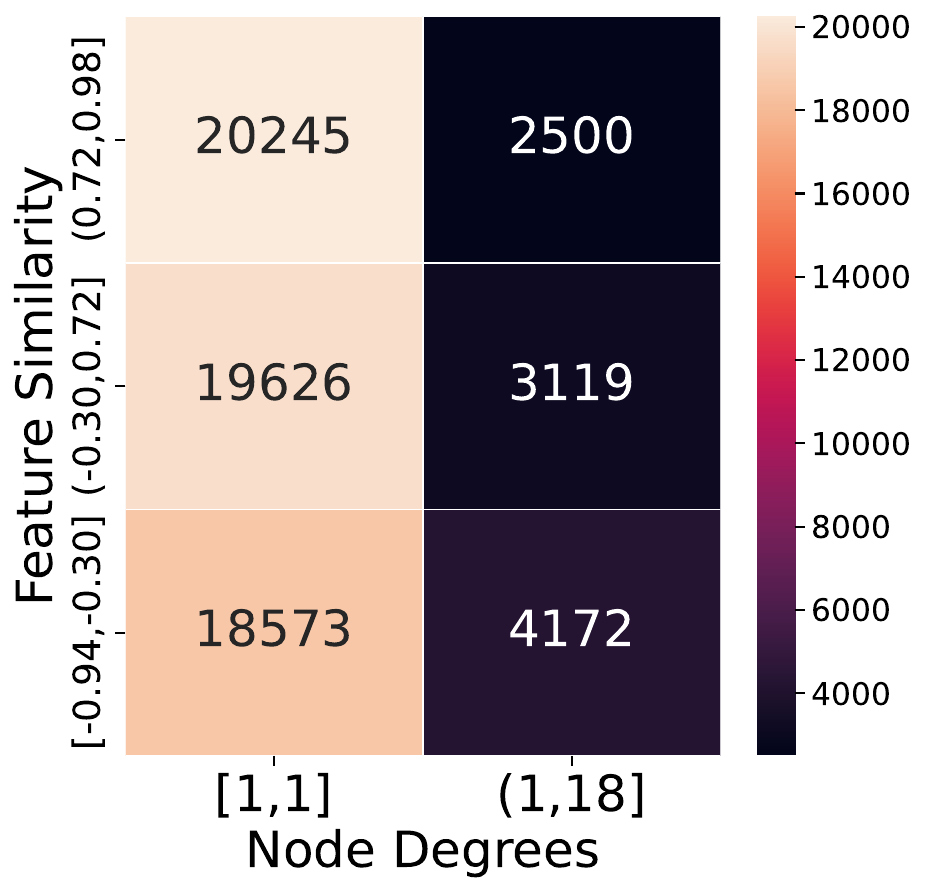}
        \caption{Number of test edges in each bucket on e-comm.}
    \end{subfigure}
    \caption{Number of edges in each bucket on different datasets.}
    \label{fig:per-edge-analysis-count}
\end{figure}

\begin{figure}
    \begin{subfigure}{0.23\textwidth}
        \includegraphics[width=\textwidth]{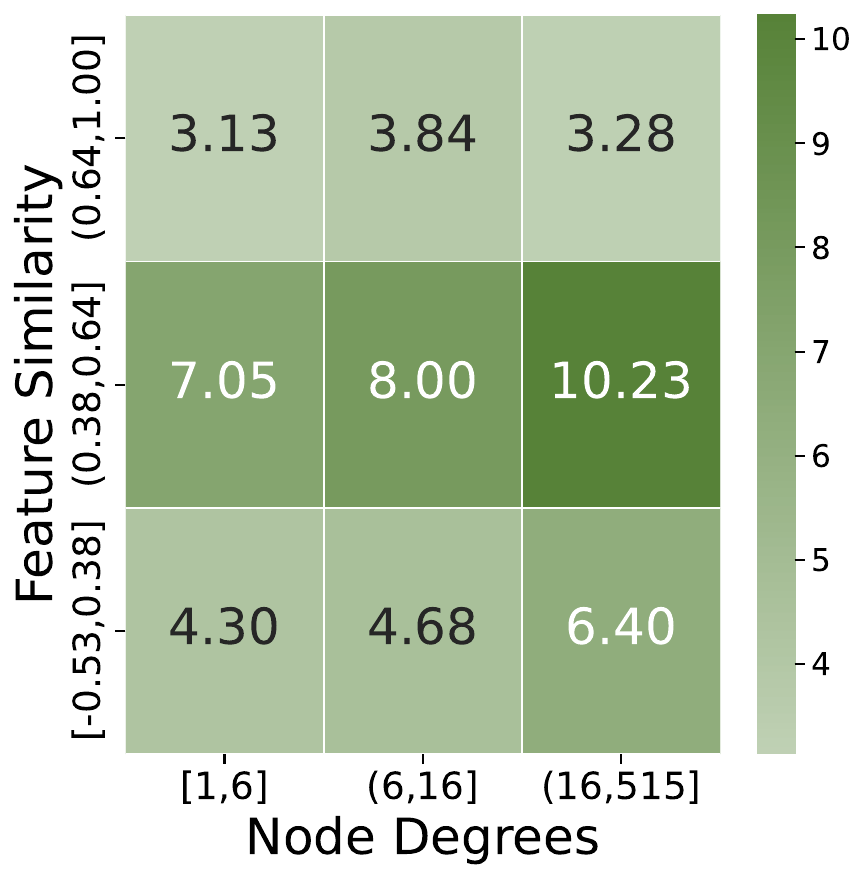}
        \caption{$\mathtt{hit}_{\mathrm{MLP}} - \mathtt{hit}_{\mathrm{DistM}}$ with SAGE encoder.}
    \end{subfigure}
    ~
    \begin{subfigure}{0.23\textwidth}
        \includegraphics[width=\textwidth]{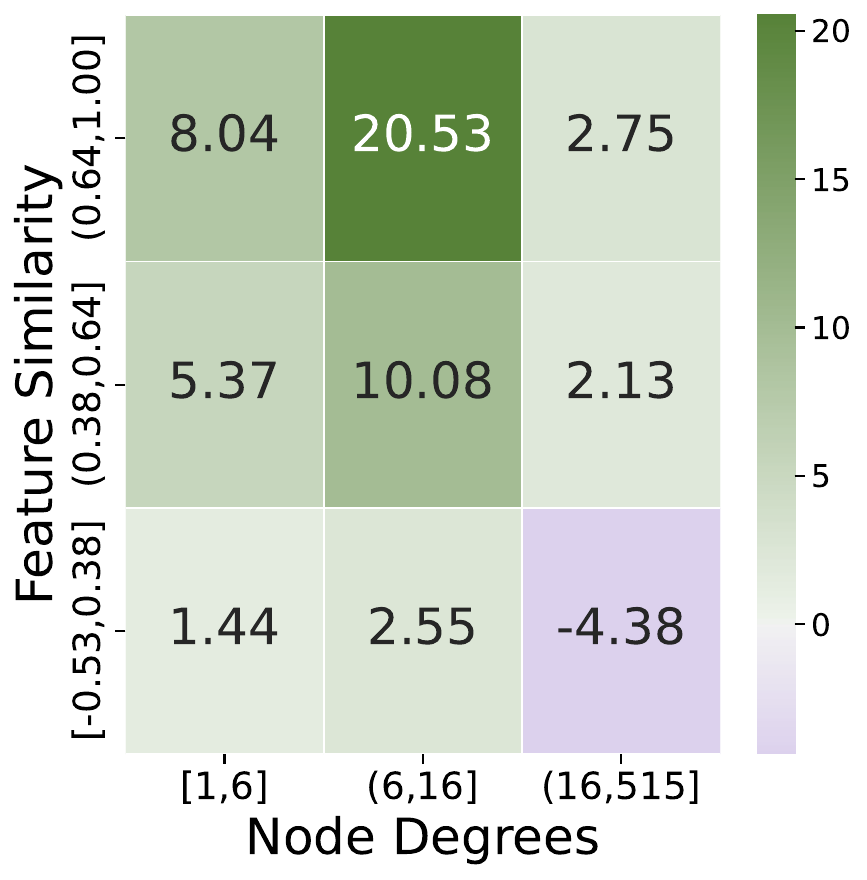}
        \caption{$\mathtt{hit}_{\mathrm{MLP}} - \mathtt{hit}_{\mathrm{DistM}}$ with GCN encoder.}
    \end{subfigure}
    ~
    \begin{subfigure}{0.23\textwidth}
        \includegraphics[width=\textwidth]{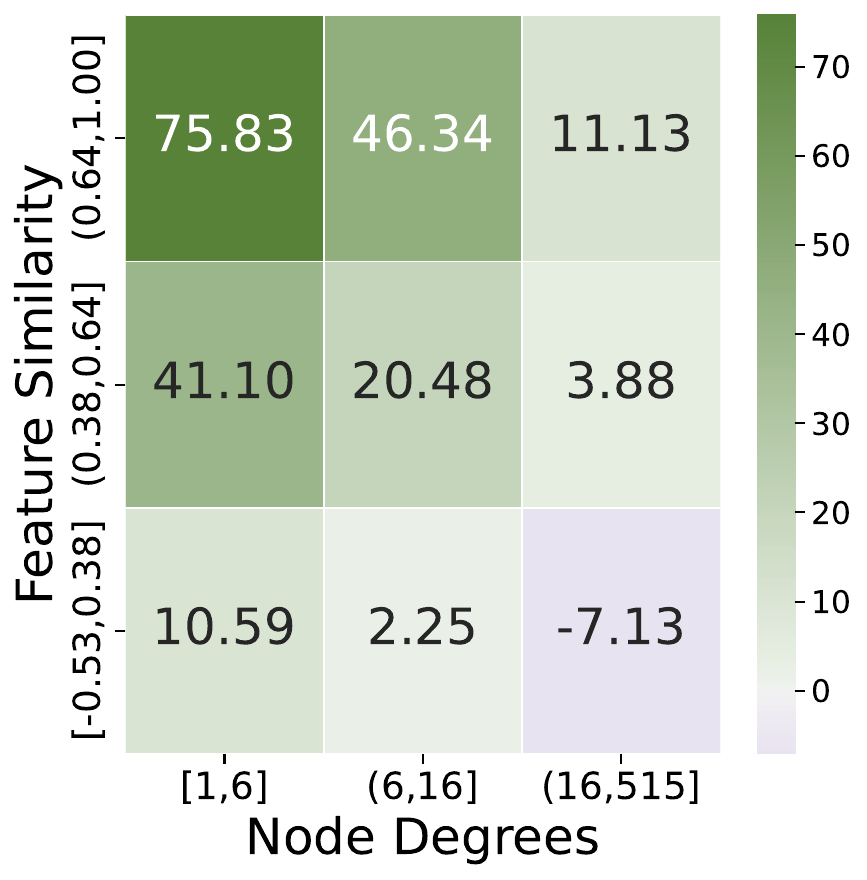}
        \caption{$\mathtt{hit}_{\mathrm{SAGE}} - \mathtt{hit}_{\mathrm{GCN}}$ with MLP decoder.}
    \end{subfigure}
    ~
    \begin{subfigure}{0.23\textwidth}
        \includegraphics[width=\textwidth]{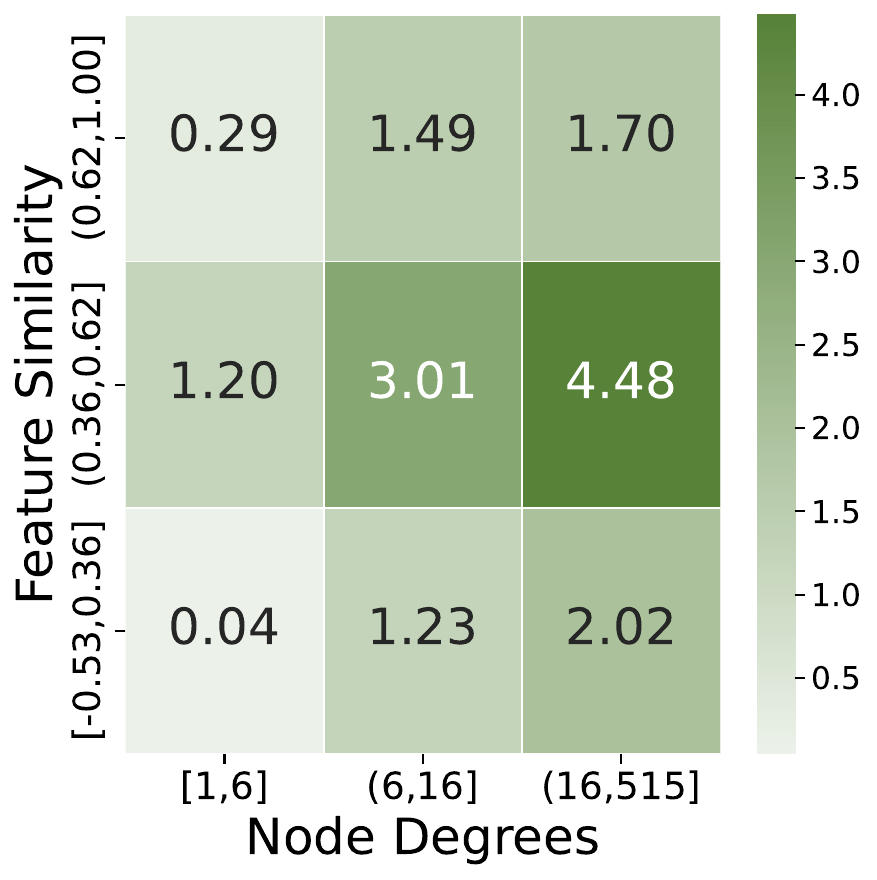}
        \caption{$\mathtt{hit}_{\mathrm{SIGN}} - \mathtt{hit}_{\mathrm{GCN}}$ with BUDDY \& MLP.}
    \end{subfigure}
    \caption{Performance comparison on ogbl-collab among different node degree and edge similarity scores buckets.}
    \label{fig:per-edge-analysis-collab}
\end{figure}

\begin{figure}
    \begin{subfigure}{0.23\textwidth}
        \includegraphics[width=\textwidth]{FIG/citation2_sage_mlp_dist.pdf}
        \caption{$\mathtt{mrr}_{\mathrm{MLP}} - \mathtt{mrr}_{\mathrm{DistM}}$ with SAGE encoder.}
    \end{subfigure}
    ~
    \begin{subfigure}{0.23\textwidth}
        \includegraphics[width=\textwidth]{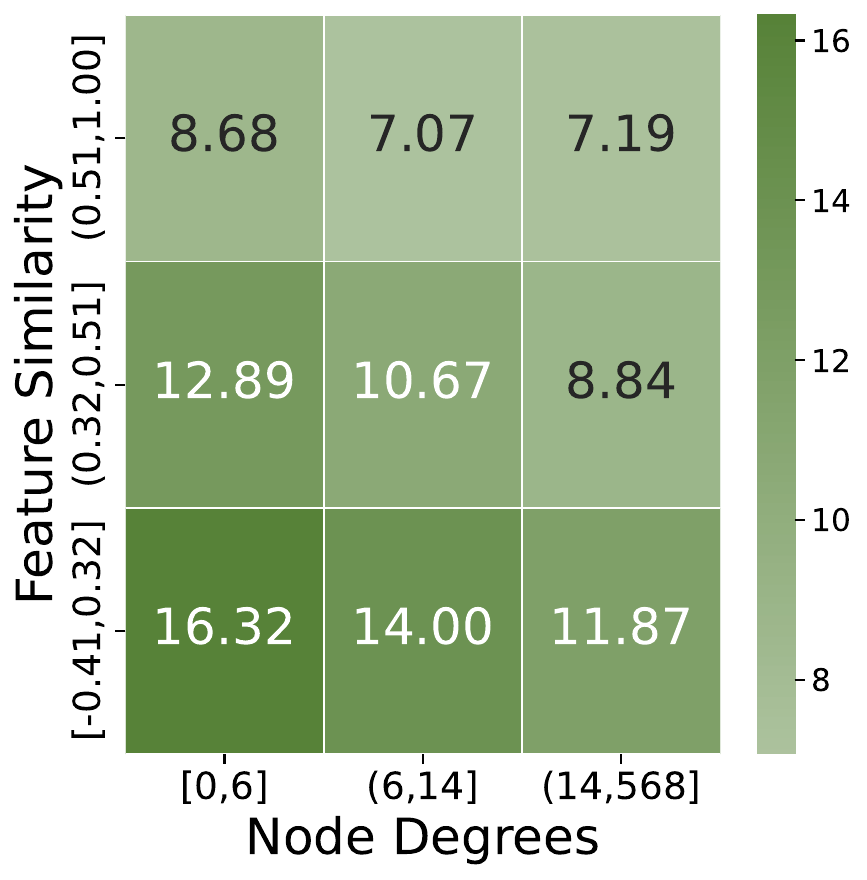}
        \caption{{$\mathtt{mrr}_{\mathrm{DistM}} - \mathtt{mrr}_{\mathrm{DOT}}$ with SAGE encoder.}}
    \end{subfigure}
    ~
    \begin{subfigure}{0.23\textwidth}
        \includegraphics[width=\textwidth]{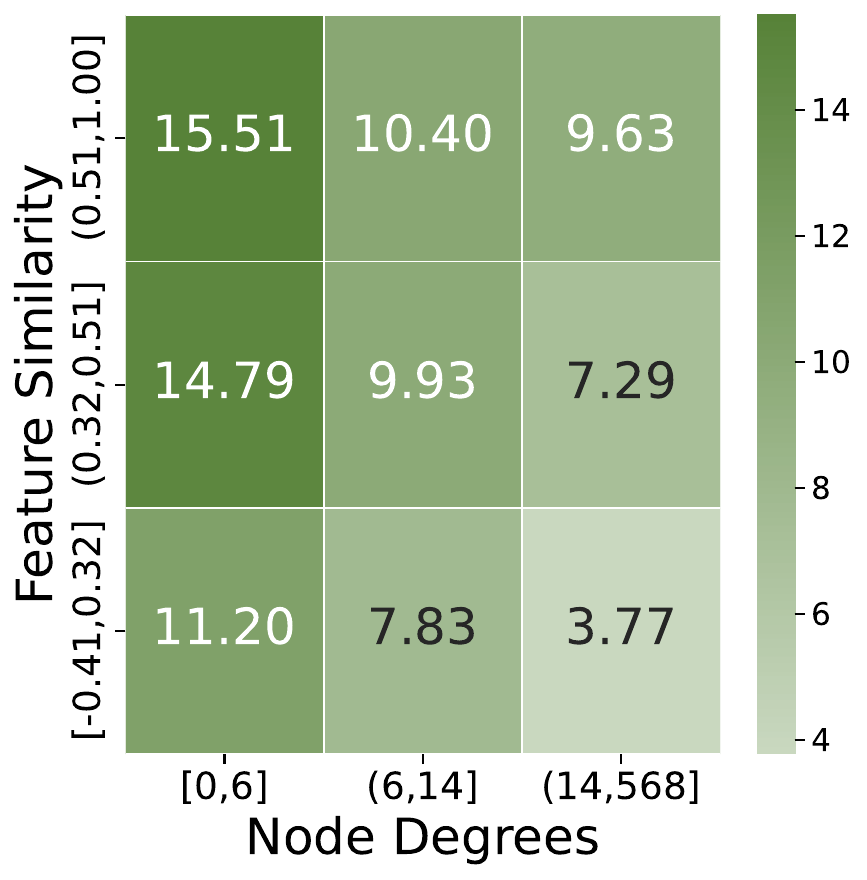}
        \caption{$\mathtt{mrr}_{\mathrm{SAGE}} - \mathtt{mrr}_{\mathrm{GCN}}$ with MLP decoder.}
    \end{subfigure}
    ~
    \begin{subfigure}{0.23\textwidth}
        \includegraphics[width=\textwidth]{FIG/citation2_buddy_sign_gcn.pdf}
        \caption{$\mathtt{mrr}_{\mathrm{SIGN}} - \mathtt{mrr}_{\mathrm{GCN}}$ with BUDDY \& MLP.}
    \end{subfigure}
    \caption{Performance comparison on ogbl-citation2 among different node degree and edge similarity scores buckets.}
    \label{fig:per-edge-analysis-citation2}
\end{figure}

\begin{figure}
    \begin{subfigure}{0.23\textwidth}
        \includegraphics[width=\textwidth]{FIG/esci_sage_mlp_dot.pdf}
        \caption{$\mathtt{mrr}_{\mathrm{MLP}} - \mathtt{mrr}_{\mathrm{DOT}}$ with SAGE encoder.}
    \end{subfigure}
    ~
    \begin{subfigure}{0.23\textwidth}
        \includegraphics[width=\textwidth]{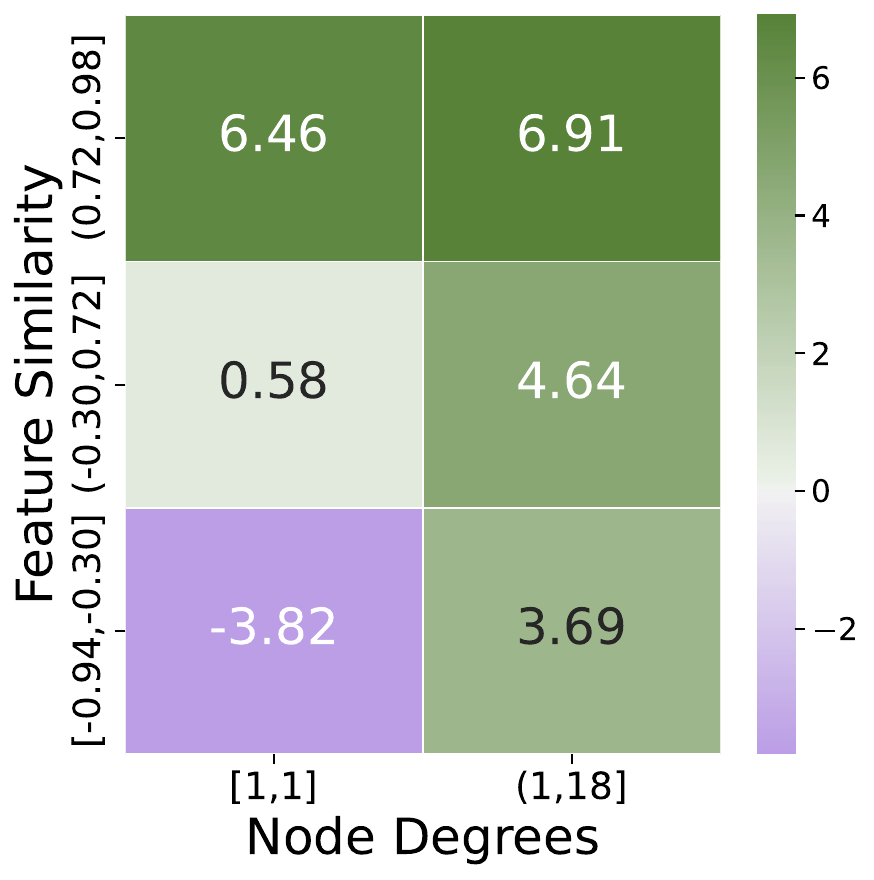}
        \caption{{$\mathtt{mrr}_{\mathrm{MLP}} - \mathtt{mrr}_{\mathrm{DistM}}$ with GCN encoder.}}
    \end{subfigure}
    ~
    \begin{subfigure}{0.23\textwidth}

        \includegraphics[width=\textwidth]{FIG/esci_sage_gcn_mlp.pdf}
        \caption{$\mathtt{mrr}_{\mathrm{SAGE}} - \mathtt{mrr}_{\mathrm{GCN}}$ with MLP decoder.}
    \end{subfigure}
    ~
    \begin{subfigure}{0.23\textwidth}
        \includegraphics[width=\textwidth]{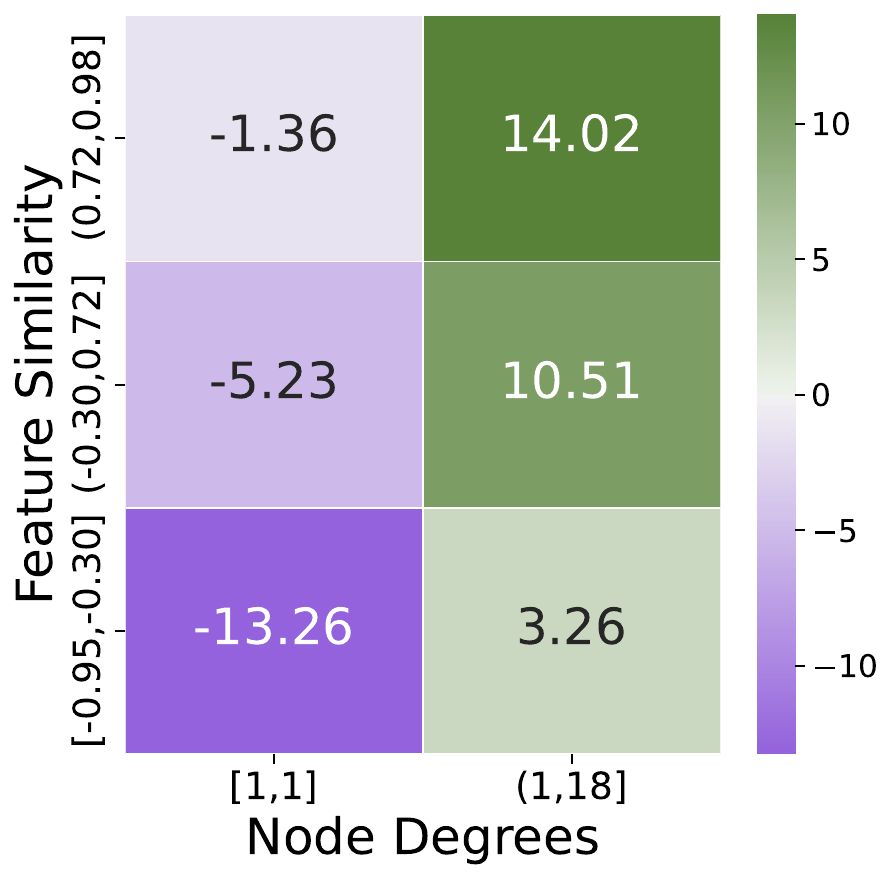}
        \caption{$\mathtt{mrr}_{\mathrm{SIGN}} - \mathtt{mrr}_{\mathrm{GCN}}$ with BUDDY \& MLP.}
    \end{subfigure}
    \caption{Performance comparison on e-comm among different node degree and edge similarity scores buckets.}
    \label{fig:per-edge-analysis-esci}
\end{figure}

\newpage
\section{Proofs of Theorems}

\subsection{Proof of Theorem~\ref{thm:hete-homo-sim-scores}}
\label{app:sec:proof-hete-homo-sim-scores}

\begin{proof}
\label{prf:hete-homo-sim-scores}

We prove separately for homophilic and heterophilic link prediction problems. 

\paragraph{Homophilic Link Prediction Problem} 
For homophilic link prediction problem, we have $1 \geq \inf(\mathcal{K}_{pos}) = M > \sup(\mathcal{K}_{neg})$. As $M = \inf(\mathcal{K}_{pos})$, a pair of nodes $u, v \in \vertexSet$ must exist in the training graph with feature vectors $\mathbf{x}_u = (\cos \theta_u, \sin \theta_u )$ and $\mathbf{x}_v = (\cos \theta_v, \sin \theta_v)$ such that $k(u, v) = \cos(\theta_u-\theta_v) = M$.

Now let us consider a fully optimized DistMult decoder for the problem. We parameterize the DistMult decoder with $\mathbf{w} = (w_1, w_2)$ and $b$; in this case, the predicted link probability 
\[ 
    \hat{y}_{uv} = (\mathbf{x}_u \otimes \mathbf{x}_v)^\T \mathbf{w} + b = w_1 \cos \theta_u \cos \theta_v + w_2 \sin \theta_u \sin \theta_v + b
\]

As $M = \inf(\mathcal{K}_{pos})$, any pairs of nodes $(u', v')$ that has feature similarity $k(u', v')$ slightly smaller than $M$ should have $\hat{y}_{u'v'} \leq 0$. Therefore, we must have $\hat{y}_{uv} = 0$. Furthermore, since we are bounding $\hat{y}_{u'v'} = 1$ if $k(u', v') = 1$ during training, we must have $\hat{y}_{uu} = 1$ and $\hat{y}_{vv} = 1$.
Using $\hat{y}_{uv} = 0$, $\hat{y}_{uu} = 1$ and $\hat{y}_{vv} = 1$, we can obtain the solutions for $w_1$, $w_2$ and $b$ as follows:
\begin{equation*}
    w_1 = \frac{1}{1-\cos(\theta_u-\theta_v)}, 
    w_2 = \frac{1}{1-\cos(\theta_u-\theta_v)}, 
    b = \frac{1}{\cos(\theta_u-\theta_v)-1}+1
\end{equation*}
As $w_1 = w_2 = \tfrac{1}{1-M}$, we have the similarity score for arbitrary node pair $(u', v')$ as
\begin{equation*}
    \hat{y}_{u'v'} 
    = \frac{1}{1-M} \cos(\theta_{u'}-\theta_{v'}) + \frac{1}{M-1}+1 
    = \frac{1}{1-M} k(u', v') + \frac{1}{M-1}+1
\end{equation*}

We can verify that the above link prediction model is fully optimized with loss function $\mathcal{L} = 0$ as it always yield $\hat{y}_{uv} \geq 0$ for $k(u', v') \geq M$ and $\hat{y}_{uv} < 0$ for $k(u', v') < M$. 
Therefore, we show that $\hat{y}_{u'v'}$ increases with $k(u', v')$ at a linear rate of $\tfrac{1}{(1-M)}$ for homophilic link prediction problem.

\paragraph{Heterophilic Link Prediction Problem}
For heterophilic link prediction problem, we have $-1 \leq \sup(\mathcal{K}_{pos}) = M < \inf(\mathcal{K}_{neg})$. As $M = \sup(\mathcal{K}_{pos})$, a pair of nodes $u, v \in \vertexSet$ must exist in the training graph with feature vectors $\mathbf{x}_u = (\cos \theta_u, \sin \theta_u )$ and $\mathbf{x}_v = (\cos \theta_v, \sin \theta_v)$ such that $k(u, v) = \cos(\theta_u-\theta_v) = M$.

Similar to the homophilic case, as $M = \sup(\mathcal{K}_{pos})$, any pairs of nodes $(u', v')$ that has feature similarity $k(u', v')$ slightly larger than $M$ should have $\hat{y}_{u'v'} \leq 0$. Therefore, we must have $\hat{y}_{uv} = 0$. Furthermore, since we are bounding $\hat{y}_{u'v'} = 1$ if $k(u', v') = -1$ during training, for nodes $u*$ and $v*$ where $\mathbf{x}_{u*} = -\mathbf{x}_{u}$ and $\mathbf{x}_{v*} = -\mathbf{x}_{v}$, we must have $\hat{y}_{uu*} = 1$ and $\hat{y}_{vv*} = 1$. Using $\hat{y}_{uv} = 0$, $\hat{y}_{uu*} = 1$ and $\hat{y}_{vv*} = 1$, we can obtain the solutions for $w_1$, $w_2$ and $b$ as follows:
\begin{equation*}
    w_1 = -\frac{1}{1+\cos(\theta_u-\theta_v)}, 
    w_2 = -\frac{1}{1+\cos(\theta_u-\theta_v)}, 
    b = 1-\frac{1}{1+\cos(\theta_u-\theta_v)}
\end{equation*}
As $w_1 = w_2 = -\tfrac{1}{1+M}$, we have the similarity score for arbitrary node pair $(u', v')$ as
\begin{equation*}
    \hat{y}_{u'v'} 
    = -\frac{1}{1+M} \cos(\theta_{u'}-\theta_{v'}) + 1-\frac{1}{1+M}
    = -\frac{1}{1+M} k(u', v') + 1-\frac{1}{1+M}
\end{equation*}
We can similarly verify that the above link prediction model is fully optimized with loss function $\mathcal{L} = 0$ as it always yield $\hat{y}_{uv} \geq 0$ for $k(u', v') \leq M$ and $\hat{y}_{uv} < 0$ for $k(u', v') > M$. Therefore, we show that $\hat{y}_{u'v'}$ decreases with $k(u', v')$ at a linear rate of $\tfrac{1}{(1-M)}$ for heterophilic link prediction problem.
\hfill $\blacksquare$
\end{proof}

\subsection{Proof of Theorem~\ref{thm:linear-decoder-limitation}}
\label{app:sec:proof-linear-decoder-limitation}
\begin{proof}
    \label{prf:linear-decoder-limitation}
    The decision boundary for gated link prediction is non-linear since it involves multiple disjoint intervals for each class (edge and non-edge). A linear model can only create a single threshold $x_0$ to separate the classes: one region for $x \leq x_0$ and the other for $x > x_0$. Therefore, there always exists misclassified points regardless of the choice of $x_0$. 
    \hfill $\blacksquare$
\end{proof}

\subsection{Proof of Theorem~\ref{thm:ego-neighbor-separation}}
\label{app:sec:proof-ego-neighbor-separation}

\begin{proof}
We begin by considering the decoder trained with the linear GNN representation $\mathbf{r}_u$ that does not separate ego- and neighbor-embeddings in its message passing. Following our assumptions, the aggregated representation  $\mathbf{r}_1$ for training nodes with feature $\mathbf{x}_1$ is $\mathbf{r}_1 = \frac{1}{d+1} \mathbf{x}_1 + \frac{d}{d+1} \mathbf{x}_2$, and $\mathbf{r}_2$ can be obtained similarly. 

Similar to Proof~\ref{prf:hete-homo-sim-scores}, we parameterize the DistMult decoder with $\mathbf{w} = (w_1, w_2)$ and $b$. Suppose node $u$ has feature vector $\mathbf{x}_1$ and node $v$ has feature vector $\mathbf{x}_2$. As the training graph is heterophilic and $(u, v)$ is connected on the graph, we should have $\hat{y}_{uv} > 0$ in this case. Without the loss of generality, we set $\hat{y}_{uv} = 1$. Furthermore, as self-loops $(u, u)$ and $(v, v)$ are not connected in the graph, we should also have $\hat{y}_{uu} = \alpha < 0$ and $\hat{y}_{vv} = \alpha < 0$. Using $\hat{y}_{uv} = 1$, $\hat{y}_{uu} = \alpha < 0$ and $\hat{y}_{vv} = \alpha < 0$, we can obtain the solutions for $w_1$, $w_2$ and $b$ as follows:
\begin{equation*}
    w_1 = w_2 = -\frac{(\alpha -1) (d+1)^2}{(d-1)^2 (\cos (\theta_1 - \theta_2)-1)}
\end{equation*}
\begin{equation*}
    b = \frac{2 \sin (\theta_1+\theta_2) \left(-d^2+(\alpha +d (\alpha  d-2)) \cos (\theta_1 - \theta_2)+2 \alpha  d-1\right)}{(d-1)^2 (-2 \sin (\theta_1+\theta_2)+\sin (2 \theta_1)+\sin (2 \theta_2))}
\end{equation*}
Now we apply the optimized DistMult decoder on the test nodes with degree $d'$. For the test node $u'$ with feature vector $\mathbf{x}_1$, the aggregated representation $\mathbf{r}_1'$ is $\mathbf{r}_1' = \frac{1}{d'+1} \mathbf{x}_1 + \frac{d'}{d'+1} \mathbf{x}_2$. Similarly, $\mathbf{r}_2'$ can be obtained. 

Assume the test nodes $u'$ has feature vector $\mathbf{x}_1$ and node $v'$ has feature vector $\mathbf{x}_2$. Based on assumptions, $(u', v')$ should be connected in the heterophilic graph. Plugging in the optimized DistMult parameters, the predicted link probability $\hat{y}_{u'v'}$ for positive (edge) node pairs can be written as
\begin{equation*}
    \hat{y}_{u'v'} = \frac{2 \alpha \left(d-d'\right) \left(d d'-1\right)-4 d d'+\left(d'\right)^2+d^2 \left(\left(d'\right)^2+1\right)+1}{(d-1)^2 \left(d'+1\right)^2}
\end{equation*}

On the other hand, for nodes with the same feature vectors (e.g., self-loops), they should not be connected in the graph. The predicted link probability $\hat{y}_{u'u'}$ for self-loops can be written as
\begin{equation*}
    \hat{y}_{u'u'} = \frac{\alpha \left(-4 d d'+\left(d'\right)^2+d^2 \left(\left(d'\right)^2+1\right)+1\right)+2 \left(d-d'\right) \left(d d'-1\right)}{(d-1)^2 \left(d'+1\right)^2}
\end{equation*}

Thus, the separation distance between edges and non-edges for the test nodes on DistMult decoder optimized with GNN representations can be represented as
\begin{equation*}
    \Delta_\mathrm{GNN} = |\hat{y}_{u'v'} - \hat{y}_{u'u'}|.
\end{equation*}

Now let us consider the baseline DistMult decoder optimized with node features $\mathbf{x}_u$. It is straight forward to see that if the decoder is optimized such that $\hat{y}_{uv} = 1$, $\hat{y}_{uu} = \alpha < 0$ and $\hat{y}_{vv} = \alpha < 0$, we will continue to have $\hat{y}_{u'v'} = 1$ and $\hat{y}_{u'u'} = \alpha$ for the test nodes, as the decoder is graph-agnostic. Thus, the separation distance between edges and non-edges for the test nodes on the baseline DistMult decoder can be represented as $\Delta_\mathrm{baseline} = 1 - \alpha$.

The GNN-based DistMult decoder reduces the separation distance between edges and non-edges for the test nodes compared to the baseline DistMult decoder when $\Delta_\mathrm{GNN} < \Delta_\mathrm{baseline}$. Solving this inequality for integers $d$ and $d'$ under the constraints of $d \geq 0, d' \geq 0, \alpha < 0$, we obtain the solutions as $d' > 0$ if $d=0$, or $1 \leq d' < d$ if $d \geq 2$. 
\hfill $\blacksquare$
\end{proof}

\section{Additional Real-World Datasets Exhibiting Feature Heterophily}

In addition to the dataset we employ in experiments, we identified a diverse range of heterophilious datasets from other graph learning tasks (e.g. graph or node classification). Specifically, we have identified a range of biological graph datasets from the TUDataset~\cite{morris2020tudataset} that exhibit feature heterophily (these datasets are typically used for graph classification). In particular, the following datasets comprise entirely heterophilic graphs (i.e., every single graph has negative homophily ratios):

\begin{itemize}
    \item aspirin
\item benzene
\item malonaldehyde
\item naphthalene
\item salicylic\_acid
\item toluene
\item uracil
\end{itemize}

These datasets are substantial in size, as detailed below:

\begin{table}[h!]
\centering
\caption{Number of entirely heterophilic Graphs in Different Datasets (from TUDataset)}
\begin{tabular}{l r}
\toprule
\textbf{Dataset} & \textbf{Number of Graphs} \\
\midrule
aspirin        & 111,763 \\
benzene        & 527,984 \\
malonaldehyde  & 893,238 \\
naphthalene    & 226,256 \\
salicylic\_acid  & 220,232 \\
toluene        & 342,791 \\
uracil         & 133,770 \\
\bottomrule
\end{tabular}
\end{table}

In addition, some datasets contain instances that are strongly heterophilious (with homophily ratio -1.0), including bbbp, NCI1, AIDS, and QM9~\citep{wu2018moleculenet}. Note that such findings are not only limited to biological datasets. For instance, 73\% of graphs from PATTERN~\cite{dwivedi2023benchmarking} (Mathematical Modeling) have negative homophily ratios.

Furthermore, we have identified node classification benchmarks from torch-geometric.datasets that display a wide range of feature homophily ratios, many of which are more heterphilious than e-comm, the one we proposed in paper. Notably, some real-world benchmarks exhibit negative homophily ratios. We summarize our findings in \Cref{tab:heterophily_dataset_node}.

\begin{figure}[h!]

    \centering
    
    \begin{subfigure}[b]{0.3\textwidth}
        \includegraphics[width=\textwidth]{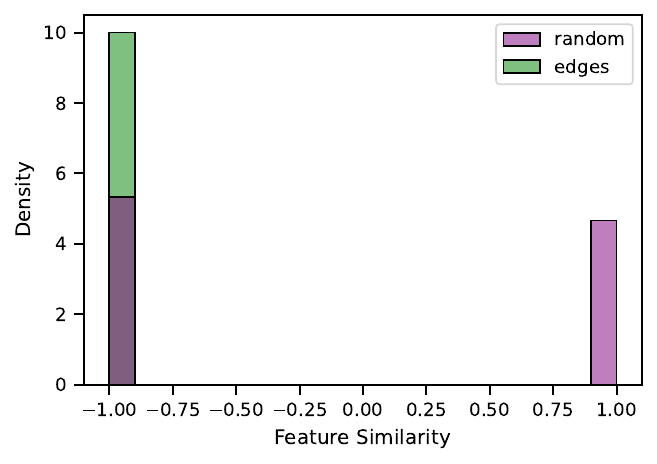}
        \caption{AIDS Instance}
    \end{subfigure}
    \hfill
    \begin{subfigure}[b]{0.3\textwidth}
        \includegraphics[width=\textwidth]{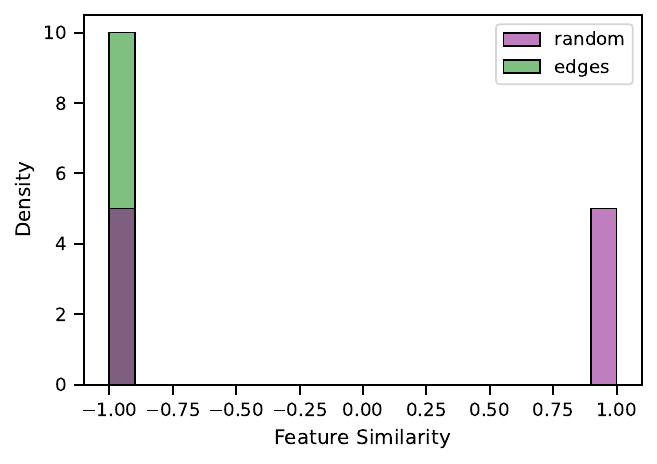}
        \caption{BBBP Instance}
    \end{subfigure}
    \hfill
    \begin{subfigure}[b]{0.3\textwidth}
        \includegraphics[width=\textwidth]{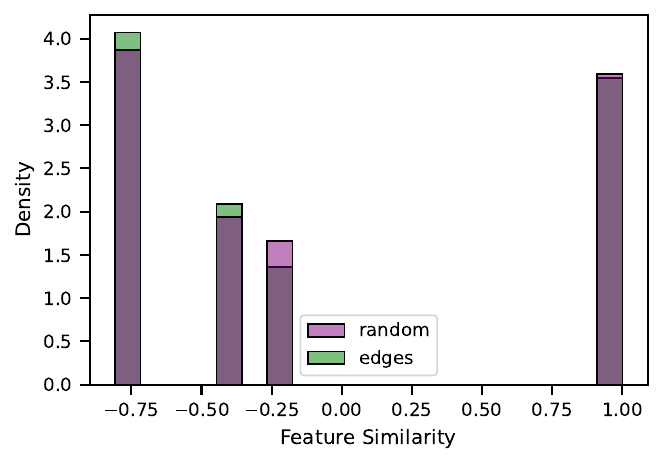}
        \caption{Beneze Instance}
    \end{subfigure}
    
    \vskip\baselineskip
    
    \begin{subfigure}[b]{0.3\textwidth}
        \includegraphics[width=\textwidth]{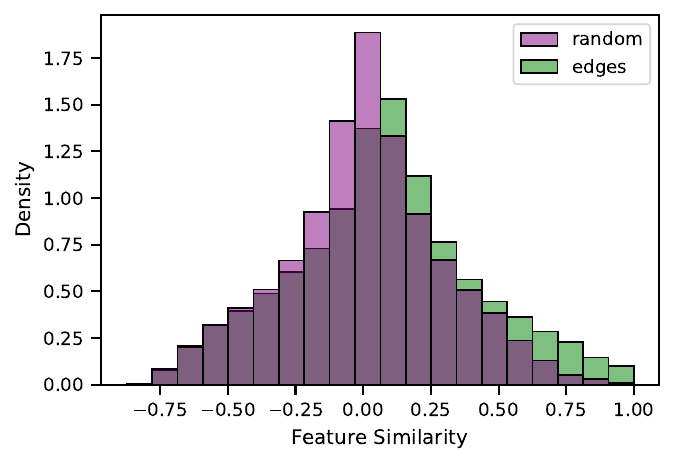}
        \caption{Computers Instance}
    \end{subfigure}
    \hfill
    \begin{subfigure}[b]{0.3\textwidth}
        \includegraphics[width=\textwidth]{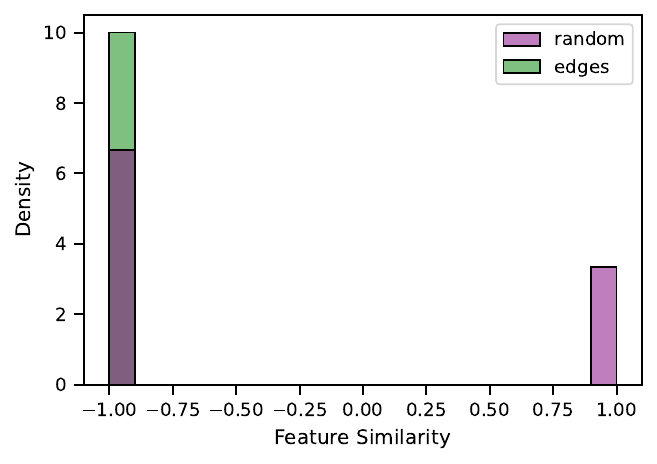}
        \caption{NC1 Instance}
    \end{subfigure}
    \hfill
    \begin{subfigure}[b]{0.3\textwidth}
        \includegraphics[width=\textwidth]{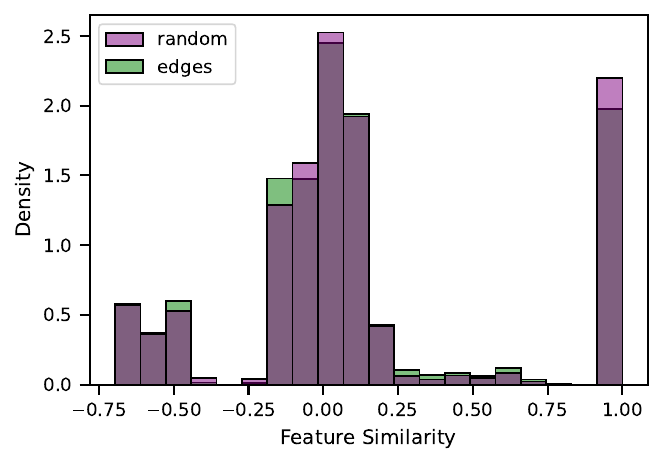}
        \caption{PPI Instance}
    \end{subfigure}
    
    \vskip\baselineskip
    
    \begin{subfigure}[b]{0.3\textwidth}
        \includegraphics[width=\textwidth]{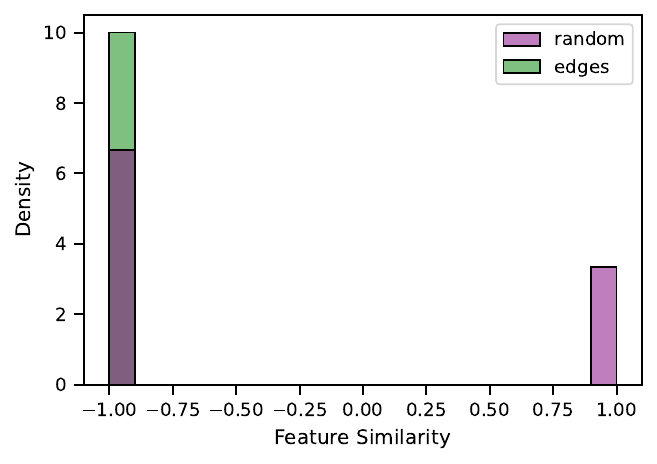}
        \caption{QM9 Instance}
    \end{subfigure}
    \hfill
    \begin{subfigure}[b]{0.3\textwidth}
        \includegraphics[width=\textwidth]{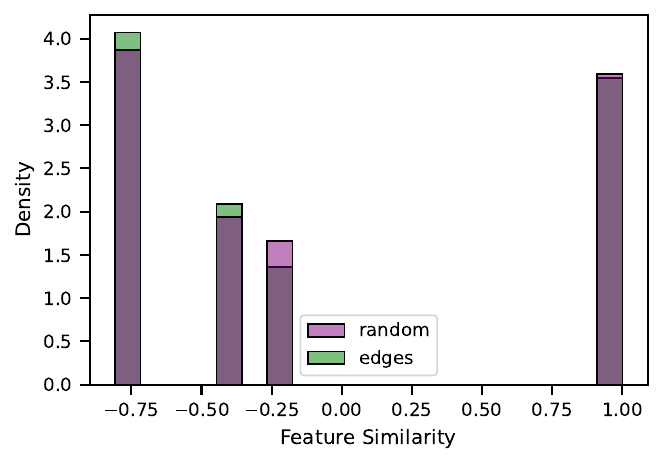}
        \caption{Salicylic Acid Instance}
    \end{subfigure}
    \hfill
    \begin{subfigure}[b]{0.3\textwidth}
        \includegraphics[width=\textwidth]{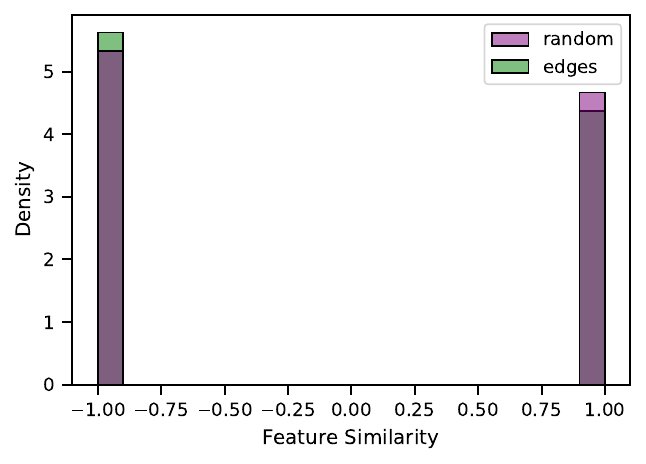}
        \caption{Toluene Instance}
    \end{subfigure}
    \caption{Heterophilic Sample Feature Distributions}

    \label{fig:heterphilic_sample_pdf}
\end{figure}

In addition to the homophily ratios presented above, we provide the feature similarity distributions for edges and random node pairs across several datasets in \Cref{fig:heterphilic_sample_pdf}, following the same convention as Figure 5 in our paper. These plots reveal clear signs of heterophily in the existing benchmark graphs.

\begin{table}[h!]
\centering
\caption{Homophily Ratios for Different Datasets}
\begin{tabular}{l r}
\toprule
\textbf{Dataset} & \textbf{Homophily Ratios} \\
\midrule
Ogbl-ppa                    & 0.74 \\
Ogbl-collab   & 0.70 \\
Ogbl-citat2   & 0.40 \\
WikiCS                      & 0.35 \\
PubMed~\cite{bojchevski2017deep}                  & 0.22 \\
e-comm         & 0.18 \\
DBLP~\cite{bojchevski2017deep}                     & 0.13 \\
Cora~\cite{bojchevski2017deep} / FacebookPagepage~\cite{rozemberczki2021multi} & 0.12 \\
AQSOL~\cite{dwivedi2023benchmarking} / Yelp             & 0.12 \\
PPI~\cite{zitnik2017predicting}                      & 0.11 \\
Facebook~\cite{yang2023pane}                 & 0.11 \\
Amazon-Photo~\cite{shchur2018pitfalls}             & 0.10 \\
Amazon-Computers~\cite{shchur2018pitfalls}        & 0.07 \\
Twitch-DE~\cite{rozemberczki2021multi}              & 0.07 \\
Twitch-FR~\cite{rozemberczki2021multi}              & 0.06 \\
BlogCatalog~\cite{yang2023pane}              & 0.06 \\
CiteSeer~\cite{yang2023pane}                 & 0.05 \\
TWeibo~\cite{yang2023pane}                   & 0.01 \\
Karateclub~\cite{zachary1977information}              & -0.03 \\
UPFD~\cite{dou2021user}                    & -0.10 \\
BBBP instances~\citep{wu2018moleculenet}           & -1.00 \\
NC11 instances~\cite{morris2020tudataset}           & -1.00 \\
AIDS instances~\cite{morris2020tudataset}           & -1.00 \\
QM9 instances~\citep{wu2018moleculenet}            & -1.00 \\
\bottomrule
\end{tabular}
\label{tab:heterophily_dataset_node}
\end{table}

\end{document}